\documentclass[10pt,twocolumn,letterpaper]{article}
\newtheorem{theorem}{Theorem}
\newtheorem{lemma}{Lemma}
\usepackage[pagenumbers]{cvpr} 

\definecolor{cvprblue}{rgb}{0.21,0.49,0.74}
\usepackage[pagebackref,breaklinks,colorlinks,allcolors=cvprblue]{hyperref}
\usepackage{multirow}
\usepackage{booktabs}
\def\paperID{} 
\def\confName{CVPR}
\def\confYear{2026}

\title{Gaussian Splatting-based Low-rank Tensor Representation  \\
for Multi-Dimensional Image Recovery}

\author{
Yiming Zeng$^{1}$ \quad Xi-Le Zhao$^{1*}$ \quad Wei-Hao Wu$^{1}$ \quad Teng-Yu Ji$^{2}$ \quad Chao Wang$^{3}$\\
$^{1}$University of Electronic Science and Technology of China, Chengdu, China\\
$^{2}$Northwest Polytechnical University, Xi’an, China\\
$^{3}$Southern University of Science and Technology, Shenzhen, China\\
{\tt\small \{yimingz20, xlzhao122003, weihaowu99\}@163.com; 
tengyu.ji@nwpu.edu.cn; wangc6@sustech.edu.cn}
}

\begin{document}
\maketitle
\begin{abstract}
     Tensor singular value decomposition (t-SVD) is a promising tool for multi-dimensional image representation, which decomposes a multi-dimensional image into a latent tensor and an accompanying transform matrix. However, two critical limitations of t-SVD methods persist: (1) the approximation of the latent tensor (\eg, tensor factorizations) is coarse and fails to accurately capture spatial local high-frequency information; (2) The transform matrix is composed of fixed basis atoms (\eg, complex exponential atoms in DFT and cosine atoms in DCT) and cannot precisely capture local high-frequency information along the mode-3 fibers. To address these two limitations, we propose a  Gaussian Splatting-based Low-rank tensor Representation (GSLR) framework, which compactly and continuously represents multi-dimensional images.  Specifically, we leverage tailored 2D Gaussian splatting and 1D Gaussian splatting to generate the latent tensor and transform matrix, respectively. The 2D and 1D Gaussian splatting are indispensable and complementary under this representation framework, which enjoys a powerful representation capability, especially for local high-frequency information. To evaluate the representation ability of the proposed GSLR, we develop an unsupervised GSLR-based  multi-dimensional image recovery model. Extensive experiments on multi-dimensional image recovery demonstrate that GSLR consistently outperforms state-of-the-art methods, particularly in capturing local high-frequency information. 
\end{abstract}
    
\section{Introduction}
\label{sec:intro}

With the rapid advancement of multi-dimensional imaging technologies,  multi-dimensional images are widely used in real-world applications such as remote sensing \cite{HAD,chen2019nonlocal}, medical imaging \cite{Medical1,Medical2}, and agricultural monitoring \cite{agriculture1, agriculture2}. Consequently, how to faithfully and effectively represent multi-dimensional images has become a fundamental research topic in computer vision.

\begin{figure}[!t]
	\centering
      \includegraphics[width=\columnwidth]{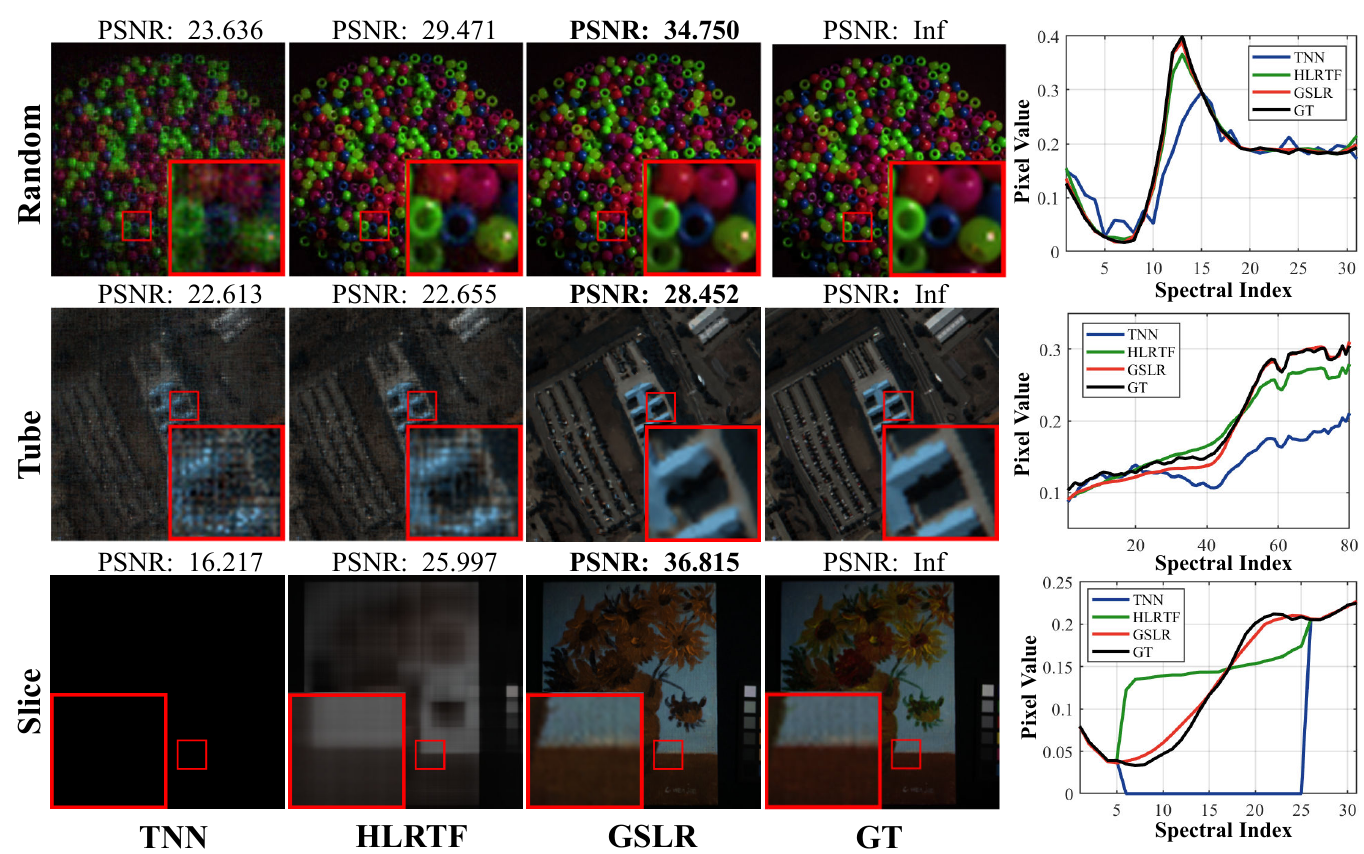}
\caption{\textbf{Comparison of different t-SVD methods for multi-dimensional image recovery under three missing patterns.}
The first four columns display the pseudo-color images reconstructed by TNN~\cite{TNN}, HLRTF~\cite{HLRTF}, the proposed GSLR, and the ground truth (GT), respectively, while the last column shows the recovered spectral curves at the spatial location $(180, 130)$.
The results clearly demonstrate that GSLR produces reconstructions most consistent with the GT, effectively recovering local high-frequency details in both the spatial dimensions and the mode-3 fibers.}
\label{fig:compare1}
\end{figure}

Recently, low-rank tensor representations have been extensively studied for multi-dimensional image representation. This is motivated by the fact that multi-dimensional images inherently exhibit strong global correlations, which can be described as low-rankness \cite{low-rank1, low-rank2, low-rank3}. Unlike matrices, the rank definition of tensors is not unique. The classic tensor ranks include the CANDECOMP/PARAFAC rank, which is defined as the minimum number of rank-one tensors required for decomposition \cite{cp1,cp2}, and the Tucker rank, which is defined by the ranks of the unfolding matrices \cite{tucker2,tucker1}. More recently, the tensor tubal-rank, which is based on the tensor singular value decomposition (t-SVD), has attracted significant attention due to the beautiful algebraic property \cite{tnn1,tnn2, TNN, tsvd3}.
Kilmer et al. \cite{kilmer2011factorization} proposed the t-SVD method, which allows the decomposition of a third-order tensor into the tensor product of two orthogonal tensors and an $f$-diagonal tensor. Based on the t-SVD, a definition of tensor rank, termed tensor tubal-rank, is introduced \cite{zhang2016exact}. This rank is numerically defined as the number of non-zero tubes in the $f$-diagonal tensor resulting from the t-SVD. This paper specifically focuses on the t-SVD.

The approximation of the latent tensor and accompanying transform matrix are crucial to the t-SVD methods. In the literature, previous methods usually employ tensor factorization to approximate the latent tensor, such as singular value decomposition (SVD) \cite{kilmer2011factorization}, nonnegative matrix factorization (NMF) \cite{HLRTF, tsvd3}, and QR decomposition \cite{QR1,QR2}. However, due to their limited representation capacity, tensor factorizations provide only a global coarse approximation of the latent tensor and fail to capture spatial local high-frequency information (\eg, sharp edges and fine textures). 
In addition, the choice of transform matrix is often restricted to classical transforms, such as the discrete Fourier transform (DFT) \cite{kilmer2011factorization,DFT} and the discrete cosine transform (DCT) \cite{DCT}. The elements of these transform matrices rely on fixed basis atoms (\eg, complex exponential atoms in DFT and cosine atoms in DCT), which cannot accurately capture local high-frequency information along the mode-3 fibers, such as disrupted spectral curves in multispectral images (MSIs).

To tackle these two challenges, we first introduce Gaussian splatting into t-SVD and propose a novel \underline{\textbf{G}}aussian \underline{\textbf{S}}platting-based \underline{\textbf{L}}ow-rank tensor \underline{\textbf{R}}epresentation (GSLR) framework for multi-dimensional images. 
Gaussian splatting models data as a weighted mixture of continuous Gaussian primitives, offering a non-neural yet effective method for modeling complex data distributions \cite{3dGS, GMM}. 
Specifically, we employ tailored 2D Gaussian splatting and 1D Gaussian splatting to generate the latent tensor and accompanying transform matrix, respectively. 
Benefiting from the powerful representation capability of tailored Gaussian splatting, the latent tensor and accompanying transform matrix achieve superior performance over previous approaches, particularly in capturing local high-frequency information. The reconstruction results of GSLR compared with previous t-SVD-based methods are shown in \cref{fig:compare1}. It can be observed that GSLR consistently outperforms existing approaches across the three representative missing patterns, particularly in capturing local high-frequency details in both the spatial dimensions (e.g., boundaries of beads and structural details of rooftops) and the mode-3 fibers (e.g., spectral curves).
Overall, our contributions are as follows:
\begin{itemize}
    \item To tackle the limitations of previous t-SVD methods in capturing local high-frequency information, we propose a novel GSLR framework for compact and continuous multi-dimensional image representation. The proposed GSLR leverages tailored 2D and 1D Gaussian splatting to faithfully capture the high-frequency information in the spatial dimensions and mode-3 fibers. Both 2D and 1D Gaussian splatting are indispensable and jointly contribute to the powerful representation capability of GSLR.
    
    \item Extensive experiments on multi-dimensional image recovery under three representative missing patterns (\eg, random missing, tube missing, and slice missing) demonstrate that the GSLR-based multi-dimensional image recovery model achieves state-of-the-art performance across diverse datasets. These results validate the powerful image representation ability of the proposed GSLR, particularly in capturing high-frequency information.
\end{itemize}

\section{Related Work}

\subsection{T-SVD Methods}
Here, we systematically review t-SVD methods from the perspective of the approximation the latent tensor and accompanying transform matrix.

The approximation of the latent tensor plays a crucial role in capturing the spatial structural information of the multi-dimensional images. In the early methods, the SVD is employed to approximate the latent tensor \cite{kilmer2011factorization}. Zhou et al. \cite{tsvd3} introduced the NMF as an alternative approach to efficiently approximate the low-rankness of the latent tensor. To further enhance accuracy and computational efficiency, Wu et al. \cite{QR1} proposed using QR decomposition as an approximation to the latent tensor, which effectively reduces computational cost.

The transform matrix is an indispensable component in capturing the information along the mode-3 fibers of a multi-dimensional image. Kilmer et al. \cite{kilmer2011factorization} first employed DFT as the transform matrix, which is composed of orthogonal complex exponential basis atoms. Madathil et al. \cite{DCT} employed orthogonal cosine basis atoms to form a real-valued DCT, achieving a more compact representation and lower computational cost than the DFT. Furthermore, Lu et al. \cite{lu2019low} required the transform matrix constructed from predefined basis atoms to be invertible, thereby deriving a new definition of tensor tubal rank. Unlike the above transforms rely on fixed basis atoms, Song et al. \cite{song2020robust} adopted data-dependent unitary basis atoms to construct a unitary transform matrix. Moreover, Luo et al. \cite{luo2022self} leveraged a fully connected neural network to implicitly learn the nonlinear basis atoms, which constructs a nonlinear transform.

Despite the success of t-SVD methods, two challenges remain: \textit{(i)}  
The approximation of the latent tensor limits the ability to capture spatial local high-frequency information, since tensor factorization provides only a coarse global representation. \textit{(ii)} The approximation of the transform matrix fails to accurately capture local high-frequency information along the mode-3 fibers, as it relies on fixed basis atoms. Although some methods employ neural networks \cite{luo2022self,HLRTF,FLRTF} to learn these basis atoms, they exhibit \textit{spectral bias} (\ie, preferentially learning low-frequency information) \cite{spectralbias}, making it difficult to precisely capture high-frequency information.

\subsection{Gaussian Splatting}
Gaussian splatting \cite{3dGS} models a scene using a set of anisotropic Gaussian primitives, which can be efficiently rasterized via $\alpha$-blending. Each Gaussian primitive is parameterized by position, opacity, covariance, and color, enabling compact memory usage while preserving the high-quality rendering for fine geometric details \cite{3DGSsurvey}. With the rapid advancement of Gaussian splatting, a wide range of applications has emerged, including scene semantic segmentation \cite{LangSplat} and generation \cite{zhou2024dreamscene360,li2024controllable}, dynamic scene reconstruction \cite{4DGS,yan2025instant}, and surgical scene modeling \cite{chen2025surgicalgs}.

Recently, Gaussian splatting has also been employed for 2D image and video modeling, 
Zhang et al. \cite{GaussianImage} proposed a novel paradigm for image representation and compression based on 2D Gaussian splatting. Building on this foundation, a number of studies have extended 2D Gaussian splatting to various image and video tasks. 
For instance, by leveraging its spatial continuity, 2D Gaussian splatting has been effectively applied to arbitrary-scale image super-resolution \cite{GaussianSR,Pixel_image}.  Liu et al. \cite{D2GV} proposed a video representation framework that integrates 2D Gaussian splatting with deformation modeling, enabling efficient and high-fidelity video reconstruction and regression. Additionally, Li et al. \cite{wangchao1} proposed a 2D Gaussian splatting-based method for hyperspectral images compression, which achieves a superior compression performance while preserving  representation quality. 

Gaussian splatting is a powerful technology for continuous modeling. However, directly adapting it to model the multi-dimensional image is non-trivial, as conventional Gaussian splatting ignores the inherent structural characteristics (\eg, low-rank structure).

\begin{figure*}[t]
     \centering   
    \includegraphics[width=.8\textwidth]{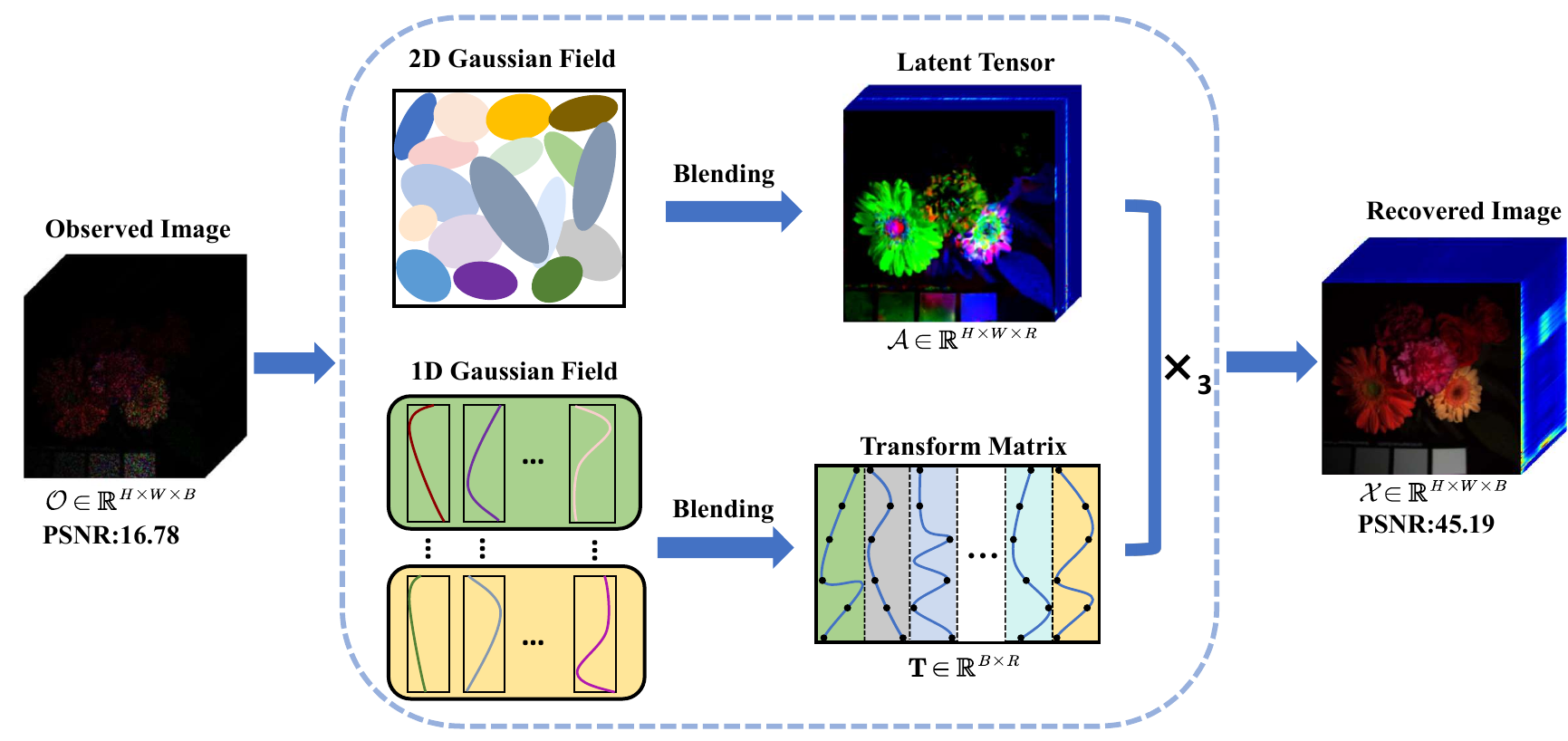}
    \caption{\textbf{Overall framework of the GSLR-based multi-dimensional image recovery model.} The input is the incomplete observed image $\mathcal{O} \in \mathbb{R}^{H\times W\times B}$. GSLR employs tailored 2D Gaussian splatting and 1D Gaussian splatting to generate the latent tensor $\mathcal{A}\in \mathbb{R}^{H\times W\times R}$ and accompanying transform matrix $\mathbf{T}\in \mathbb{R}^{B\times R}$, respectively. By applying the transform $\mathbf{T}$ to the latent tensor $\mathcal{A}$, we obtain the recovered image $\mathcal{X} \in \mathbb{R}^{H\times W\times B}$.}
    \label{fig:framework}
\end{figure*}

\section{Notations and Preliminaries}
In this paper, scalars, vectors, matrices, and tensors are denoted by lowercase letters (\eg, $x$), bold lowercase letters (\eg, $\mathbf{x}$), bold uppercase letters (\eg, $\mathbf{X}$), and calligraphic uppercase letters (\eg, $\mathcal{X}$), respectively. For a third-order tensor $\mathcal{X} \in \mathbb{R}^{H \times W \times B}$, we use $\mathcal{X}_{[i]}$ to denote the $i$-th frontal slice $\mathcal{X}(:,:,i) \in \mathbb{R}^{H \times W}$. The $(i,j,k)$-th element of $\mathcal{X}$ is denoted by $\mathcal{X}(i,j,k)$, and the Frobenius norm is defined as $\|\mathcal{X}\|_F = \sqrt{ \sum_{i,j,k} |\mathcal{X}(i,j,k)|^2 }$. The matrix nuclear norm of $\mathbf{X} \in \mathbb{R}^{H \times W}$ is defined as $\|\mathbf{X}\|_* = \sum_{i=1}^{\min(H, W)} \sigma_i(\mathbf{X})$, where $\sigma_i(\mathbf{X})$ is the $i$-th singular value of $\mathbf{X}$.
For a tensor $\mathcal{X} \in \mathbb{R}^{H \times W \times B}$, the unfolding operator is defined as $\text{unfold}_3(\cdot): \mathbb{R}^{H \times W \times B} \to \mathbb{R}^{B \times HW}$, with the resulting matrix denoted as $\mathbf{X}^{(3)} = \text{unfold}_3(\mathcal{X})$. Its inverse operation is denoted by $\text{fold}_3(\cdot)$. The mode-3 tensor-matrix product of a third-order tensor $\mathcal{X} \in \mathbb{R}^{H \times W \times R}$ and a matrix $\mathbf{T} \in \mathbb{R}^{B \times R}$ is defined as $\mathcal{Y}=\mathcal{X} \times_3 \mathbf{T} := \text{fold}_3(\mathbf{T} \mathbf{X}^{(3)}) \in \mathbb{R}^{ H \times W \times B}$. 

\section{Proposed Method}

\subsection{Proposed GSLR}
T-SVD is a promising tool for compactly representing the multi-dimensional images.
However, the approximation of the latent tensor is coarse and fails to accurately capture spatial local high-frequency information. The transform matrix is composed of fixed basis atoms and cannot precisely capture local high-frequency information along the mode-3 fibers.
Gaussian splatting, a powerful tool in 3D scene reconstruction known for its high-quality and efficiency, inspires a potential solution to these two limitations. 

In this paper, we propose a novel Gaussian splatting-based Low-rank tensor Representation (GSLR) for multi-dimensional images.
Specifically, the multi-dimensional image $\mathcal{X}\in \mathbb{R}^{H\times W\times B}$ is decomposed into a latent tensor $\mathcal{A}\in \mathbb{R}^{H\times W\times R}$ and an accompanying transform matrix $\mathbf{T}\in \mathbb{R}^{B\times R}$. The latent tensor is generated via tailored 2D Gaussian splatting, while the accompanying transform matrix is generated through tailored 1D Gaussian splatting. Mathematically, the proposed GSLR can be formulated as:
\begin{flalign}
    \label{eq:gslr_1}
    && \mathcal{X} &= \mathcal{A} \times_3 \mathbf{T}, && \\
    \text{where} \quad && \mathcal{A}(x,y) &= \sum_{j=1}^{N}\mathcal{G}_{\text{2D}}(x,y;\,\boldsymbol{\mu}_j,
     \mathbf{\Sigma}_j,\mathbf{c}_j), && \notag\\
    && \mathbf{T}(z, r) &= \sum_{k=1}^{K}\mathcal{G}_{\text{1D}}(z;\,\mu_k^r, \sigma_k^r, c_k^r). && \notag
\end{flalign}
Here, $\mathcal{A}(x,y)\in \mathbb{R}^R$ is the value of the latent tensor at the spatial coordinate $(x, y)$;
$\mathbf{T}(z, r)$ is the value at the spectral coordinate $z$ of the $r$-th column in the transform matrix $\mathbf{T}$;
$N$ and $K$ denote the number of 2D Gaussian primitives in 2D Gaussian splatting and 1D Gaussian primitives in 1D Gaussian splatting, respectively;
$\mathcal{G}_{\text{2D}}(x, y;, \boldsymbol{\mu}_j, \mathbf{\Sigma}_j, \mathbf{c}_j) \in \mathbb{R}^R$ is the value of the 2D Gaussian primitive at $(x, y)$, where $\boldsymbol{\mu}_j$, $\mathbf{\Sigma}_j$, and $\mathbf{c}_j$ are the learnable parameters of the 2D Gaussian primitive.
Similarly, $\mathcal{G}_{\text{1D}}(z;, \mu_k^r, \sigma_k^r, c_k^r) \in \mathbb{R}$ is the value of the 1D Gaussian primitive at $z$, where $\mu_k^r$, $\sigma_k^r$, and $c_k^r$ are the learnable parameters of the 1D Gaussian primitive.

Next, we will provide a detailed introduction to the two essential components of the proposed GSLR.

\subsection{\textbf{2D Gaussian Splatting-based Latent Tensor Approximation}}
To accurately capture the spatial structural information of the multi-dimensional image, particularly spatial local high-frequency information, we employ a tailored 2D Gaussian splatting to generate the latent tensor. Essentially, the latent tensor is compactly parameterized as a continuous 2D Gaussian field, which consists of numerous continuous 2D Gaussian primitives.
Each primitive is defined by following learnable parameters: position $\boldsymbol{\mu} \in \mathbb{R}^2$, covariance $\mathbf{\Sigma} \in \mathbb{R}^{2\times 2}$, and feature $\mathbf{c}\in \mathbb{R}^R$ \cite{GaussianImage}. Here, the feature is directly defined as a vector whose number of elements (\ie, $R$) corresponds to the mode-3 dimensionality of the latent tensor.

\begin{figure}[!h]
	\centering
      \includegraphics[width=\columnwidth]{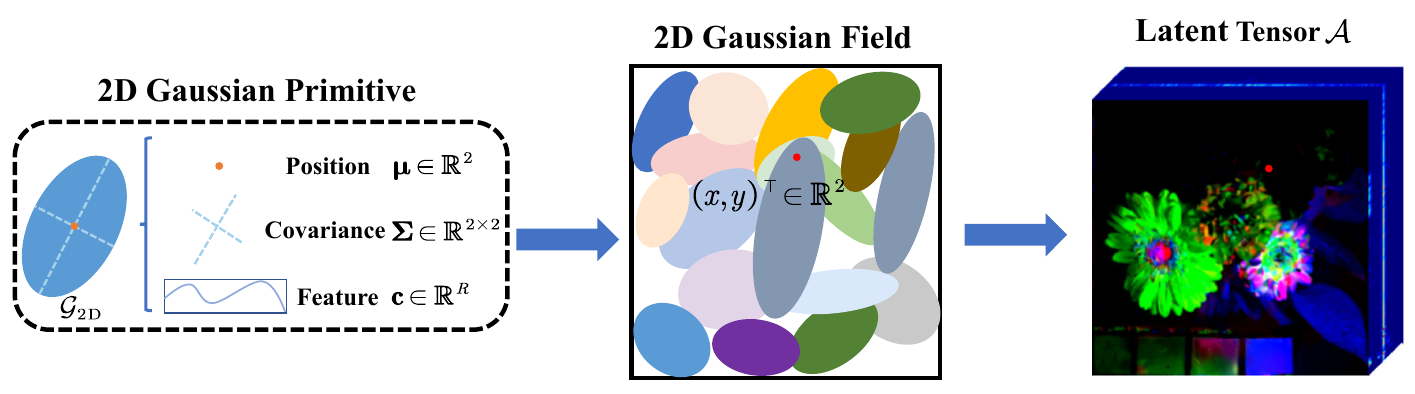}
\caption{The framework of the 2D Gaussian splatting-based latent tensor approximation.}
\label{fig:2dgs_f}
\end{figure}

The value of the latent tensor at arbitrary spatial coordinate is obtained by rendering the overlapped 2D Gaussian primitives from the continuous 2D Gaussian field, which is referred to as \textit{blending}.
Consider a continuous 2D Gaussian field composed of $N$ 2D Gaussian primitives, for an arbitrary spatial coordinate $(x, y) \in \{1,2,\dots,H\} \times \{1,2,\dots,W\} \subset \mathbb{R}^2$, the value $\mathcal{A}(x,y) \in \mathbb{R}^R$ can be rendered as follows:
\begin{equation}
    \label{eq:2dgs}
    \begin{aligned}
        \mathcal{A}(x,y) &=  \sum_{j=1}^{N} \mathcal{G}_\text{2D}(x,y;\boldsymbol{\mu}_j, \mathbf{\Sigma}_j, \mathbf{c}_j)\\
        &=\! \scalebox{0.9}{$\displaystyle \! \!\sum_{j=1}^{N} \mathbf{c}_j \! \cdot\! \exp\!\!\left(\!\!-\frac{1}{2}\! \! \left(\! \!\begin{pmatrix}\!x \\y\end{pmatrix}\! - \! \boldsymbol{\mu}_j\! \!\right)^\top \!\!\!\mathbf{\Sigma}_j^{-1}\! \! \left(\!\begin{pmatrix}x \\y\end{pmatrix}\! -  \!\boldsymbol{\mu}_j\right)\!\!\right), $}
    \end{aligned}
\end{equation} 
where $\boldsymbol{\mu}_j $, $\mathbf{\Sigma}_j$, and $\mathbf{c}_j$ represent the position, covariance, and feature of the $j$-th 2D Gaussian primitive, respectively. 
The parameters of 2D Gaussian splatting are denoted as:
\begin{equation}
    \theta_\mathcal{A} = \{\boldsymbol{\mu}_j, \mathbf{\Sigma}_j, \mathbf{c}_j \mid   j=1,2,\dots, N\}.
\end{equation}
Each 2D Gaussian primitive has ($5+R$) learnable parameters, which consist of 2 for position, 3 for covariance, and $R$ for feature. Therefore, the total number of learnable parameters for 2D Gaussian splatting is $N(5 + R)$.  The framework of the 2D Gaussian splatting-based latent tensor approximation is shown in \cref{fig:2dgs_f}.

\subsection{\textbf{1D Gaussian Splatting-based Transform Matrix Approximation}}

To batter capture the high-frequency information along the mode-3 fibers of the multi-dimensional image, we extend Gaussian splatting to a tailored 1D case and leverage the powerful representation capacity to generate the corresponding transform matrix. A detailed description of the 1D Gaussian splatting formulation is provided below.

Similar to 2D Gaussian splatting, the transform matrix is compactly parameterized as a set of 1D Gaussian fields. Specifically, we leverage $R$ continuous 1D Gaussian fields to generate the columns of the transform matrix.
Each 1D Gaussian field, which consists of $K$ continuous 1D Gaussian primitives,
is responsible for generating one column of the transform matrix. Here, each 1D Gaussian primitive is defined by three learnable parameters: position $\mu \in \mathbb{R}$, variance $\sigma \in \mathbb{R}^{+}$, and feature $c \in \mathbb{R}$.

\begin{figure}[!h]
	\centering
      \includegraphics[width=\columnwidth]{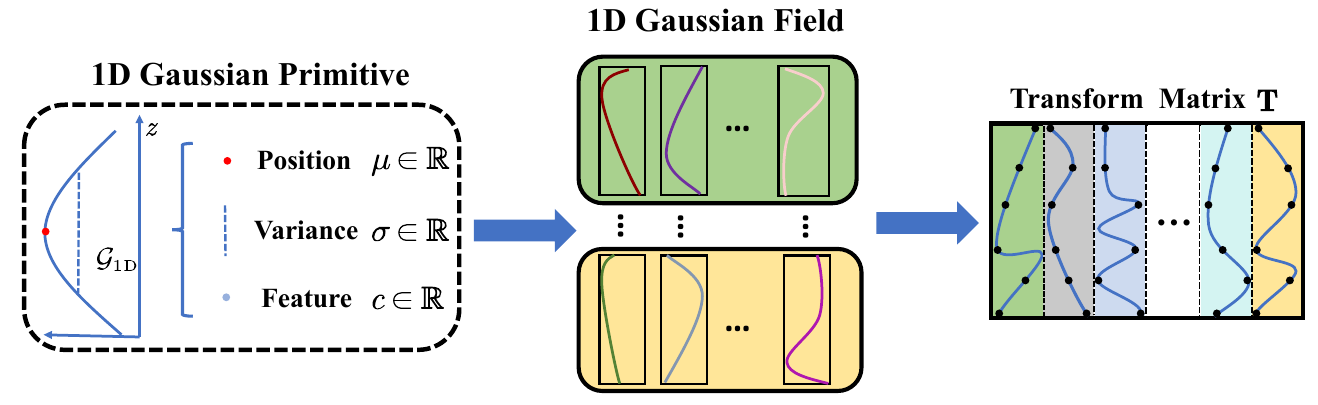}
\caption{The framework of the 1D Gaussian splatting-based transform matrix approximation.}
\label{fig:1dgs_f}
\end{figure}

The value of the transform matrix at arbitrary spectral coordinate is generated by rendering the overlapped 1D Gaussian primitives from the corresponding 1D Gaussian field, which is referred to as \textit{blending}. Specifically, for an arbitrary spectral coordinate $z \in \{1,2,\ldots, B\}$ in the $r$-th ($r\in \{1,2,\ldots, R\}$) column of the transform matrix $\mathbf{T}(z, r) \in \mathbb{R}$, can be rendered via:
\begin{equation}
    \begin{aligned}
        \mathbf{T}(z,r)&=  \sum_{k=1}^K \mathcal{G}_\text{1D}(z; \mu^r_k, \sigma^r_k,c^r_k)\\
        &=  \sum_{k=1}^K c^r_k  \cdot   \exp\left(-\frac{(z- \mu^r_k)^2}{2(\sigma^r_k)^2}\right),
    \end{aligned}
\end{equation}
where $\mu_k^r $, $\sigma_k^r$, and $c_k^r$ represent the position, variance, and feature of the $k$-th 1D Gaussian primitive from the $r$-th 1D Gaussian field, respectively. The learnable parameters of 1D Gaussian splatting are collectively denoted as:
\begin{equation}
\theta_{\mathbf{T}} = \left\{\mu^r_k,\, \sigma^r_k,\, c^r_k \mid k = 1,\dots,K;\, r = 1,\dots,R \right\}.
\end{equation}
Here, each 1D Gaussian primitive contains 3 learnable parameters. The total number of learnable parameters for 1D Gaussian splatting is $3KR$.
The framework of the 1D Gaussian splatting-based transform matrix approximation is illustrated in \cref{fig:1dgs_f}.

\subsection{Connection to Classical t-SVD Method}
The proposed GSLR introduces a novel framework for multi-dimensional images representation based on Gaussian splatting, which enjoys a strong representation capability. We now discuss the connection between the proposed GSLR and classical t-SVD method.
\begin{lemma}\label{lemma1}
	\textit{When the parameters $\theta_{\mathcal{A}}$ and $\theta_{\mathbf{T}}$ of the 2D Gaussian splatting and the 1D Gaussian splatting satisfy:
		\begin{enumerate}[label=(\roman*),leftmargin=2em]
			\item $N=HW$, $\boldsymbol{\mu}_j=(x,y)^\top$, and $j=(x-1)H+y$ for $j=1,2,\ldots,N$;
			\item $K=B$, $\mu_{k}^{\,r}=k$ for $k=1,2,\ldots,K$ and $r=1,2,\ldots,B$;
			\item $\mathbf{\Sigma}_{j}\to \mathbf{0}$ and $\sigma_{k}^{\,r}\to 0$,
		\end{enumerate}
		then the 2D and 1D Gaussian splatting can explicitly represent arbitrary latent tensor $\mathcal{A}$ and arbitrary transform matrix $\mathbf{T}$.}
\end{lemma}
\begin{theorem}\label{theorem1}
	\textit{Under the conditions of Lemma~\ref{lemma1}, when the following conditions hold:
		\begin{enumerate}[label=(\roman*),leftmargin=2em]
			\item $\mathcal{A}_{[r]}=\mathbf{U}_{[r]}\,\mathbf{S}_{[r]}\,\mathbf{V}_{[r]}^{\mathrm H}$ with unitary matrices $\mathbf{U}_{[r]},\mathbf{V}_{[r]}$ and diagonal matrix $\mathbf{S}_{[r]}$;
			\item $\mathbf{T}$ is the discrete Fourier transform matrix,
		\end{enumerate}
		 then the proposed GSLR degrades to the classical t-SVD method.}
\end{theorem}

Theorem \ref{theorem1} reveals that the proposed GSLR enjoys a more powerful representation capability than the classical t-SVD method, since GSLR can degenerate into the classical t-SVD when its Gaussian splatting satisfy certain conditions. 
The detailed proof is provided in the supplementary material.

\subsection{Unsupervised Multi-Dimensional Image Recovery Model}
To evaluate the representation ability of GSLR, we develop an unsupervised GSLR-based multi-dimensional image recovery model. Specifically, the GSLR-based recovery model can be formulated as follows:
\begin{equation}
\begin{aligned}
    \min_{\theta_{\mathcal{A}},\theta_{\mathbf{T}}}\,\,
    & \left\|\mathcal{M}\odot(\mathcal{O} - \mathcal{A} \times_3 \mathbf{T}) \right\|_F^2
    +  \lambda \sum_{i=1}^R\left\| \mathcal{A}_{[i]} \right\|_{*}, \\ 
    \text{s.t.} \,\,\,\,
    & \mathcal{A}(x,y) = \sum_{j=1}^{N} \mathcal{G}_\text{2D}(x,y;\,\boldsymbol{\mu}_j, \mathbf{\Sigma}_j,\mathbf{c}_j), \label{eq:gslr}\\
    & \mathbf{T}(z, r) = \sum_{k=1}^K  \mathcal{G}_\text{1D}(z;\, \mu_k^r, \sigma_k^r,c_k^r),
\end{aligned}
\end{equation}
where $\mathcal{O} \in \mathbb{R}^{H\times W\times B}$ is the observed multi-dimensional image, $\mathcal{M} \in \mathbb{R}^{H\times W\times B}$ is a binary mask that assigns ones to the observed entries and zeros to the unobserved ones, and $\odot$ denotes the point-wise multiplication.
In addition, we consider the spatial low-rank prior of the multi-dimensional image and impose a slice-wise matrix nuclear norm constraint on frontal slices of the latent tensor with $\lambda$ serving as the trade-off parameter.

In GSLR, all learnable parameters, $\theta_\mathcal{A}$ and $\theta_\mathbf{T}$,  with a total of $N(5+R) + 3KR$ parameters.
Since the blending processes in both 2D and 1D Gaussian splatting are fully differentiable, we directly use off-the-shelf gradient descent-based optimizer (\ie, Adam \cite{adam}) to solve problem (\ref{eq:gslr}).
Once the optimal $\theta_\mathcal{A}$ and $\theta_\mathbf{T}$ are obtained, the recovered multi-dimensional image can be represented by \cref{eq:gslr_1}. The framework of GSLR-based multi-dimensional image recovery model is shown in \cref{fig:framework}.


\section{Experiments}
\subsection{Experimental Settings}

\textbf{Compared Methods.}  We compare GSLR-based multi-dimensional image recovery model with six state-of-the-art methods, \ie, TNN \cite{TNN}, TRLRF \cite{TRLRF},
TCTV \cite{TCTV}, HLRTF \cite{HLRTF}, LRTFR \cite{TRLRF}, and FLRTF \cite{FLRTF}. To ensure a fair comparison, all parameters associated with the compared methods are carefully tuned according to the suggestions provided by the respective authors in their articles. 

In our method, the number of 2D Gaussian primitives $N$ is searched within the candidate set $\{i \times 10^{4}\}_{i=1}^9$. 
The number of 1D Gaussian primitives $K$ is explored within the interval $[20, 100]$ with a step size of 5. 
The trade-off parameter $\lambda$ is tuned over the candidate set $\{10^{-i}\}_{i=1}^5$. 
The dimensionality parameter $R$ of the latent tensor is adjusted within $\{15, 20, \dots, 55, 60\}$ for random and tube missing patterns, and within $\{100, 150, 200, 250, 300\}$ for slice missing pattern.
The learning rate of the Adam optimizer is tuned over the set $\{10^{-2}, 10^{-3}\}$. 
For both 2D and 1D Gaussian primitives, their attributes (\ie, position, covariance or variance, and feature) are randomly initialized.

\noindent\textbf{Implementation}. 
In experiments, we employ two types of multi-dimensional image: color images\footnote{[Online]. Avaialble: https://sipi.usc.edu/database/} and MSIs\footnote{[Online]. Avaialble: https://cave.cs.columbia.edu/repository/Multispectral and https://www.ehu.eus/ccwintco/index.php}.  All datasets are normalized to the range [0, 1] before experimentation. The sampling rates (SRs) are set to 0.02, 0.05, and 0.10 for random missing cases, and to 0.10, 0.15, and 0.20 for tube missing cases.
In addition, we consider a more challenging scenario: slice missing, to evaluate the model's ability to capture information along the mode-3 fibers. In multi-dimensional image acquisition, slice-wise corruption often occurs due to environmental interference or sensor malfunction, resulting in the complete loss of the mode-3 slices \cite{schowengerdt2006remote}. 
Specifically, we select five MSIs and set the first five and the last five bands as the sampling bands, while the remaining bands are treated as missing.

\noindent\textbf{Evaluation Metrics.} 
Two quantitative evaluation metrics, peak signal-to-noise ratio (PSNR) and structural similarity (SSIM), are selected to evaluate the overall quality of the recovered results. In general, the best values for PSNR and SSIM are infinity and 1, respectively. Higher PSNR and SSIM values indicate better recovery quality.

\begin{table}[!h]
  \centering
    \footnotesize
  \setlength{\tabcolsep}{3pt}
\renewcommand\arraystretch{1}
  \caption{The average quantitative results for random missing. The \textbf{best} and \underline{second-best} values are highlighted.}
    \begin{tabular}{cccccccc}
    \toprule
    \multicolumn{2}{c}{Sampling Rate} & \multicolumn{2}{c}{0.02} & \multicolumn{2}{c}{0.05} & \multicolumn{2}{c}{0.10} \\
    \midrule
    Data  & Method & \multicolumn{1}{c}{PSNR} & \multicolumn{1}{c}{SSIM} & \multicolumn{1}{c}{PSNR} & \multicolumn{1}{c}{SSIM} & \multicolumn{1}{c}{PSNR} & \multicolumn{1}{c}{SSIM} \\
    \midrule
    \multirow{7}[0]{*}{\shortstack{\textbf{Color images}\\\textit{Plane}\\
    \textit{House}\\\textit{Baboon}\\{\tiny $(512\times 512\times 3)$}\\\textit{Female}\\{\tiny $(256\times 256\times 3)$}}} & TNN   & 15.669 & 0.194 & 18.003 & 0.297 & 20.143 & 0.419 \\
          & TRLRF & 14.424 & 0.104 & 17.418 & 0.242 & 20.464 & 0.425 \\
          & TCTV  & 18.949 & \underline{0.476} & 21.811 & \underline{0.591} & \underline{24.416} & \underline{0.705} \\
          & HLRTF & 16.588 & 0.233 & 19.388 & 0.362 & 22.108 & 0.542 \\
          & LRTFR & \underline{19.195} & 0.380 & \underline{22.279} & 0.540 & 24.321 & 0.661 \\
          & FLRTF & 14.166 & 0.103 & 17.787 & 0.278 & 20.664 & 0.441 \\
          & GSLR  & \textbf{21.684} & \textbf{0.571} & \textbf{23.578} & \textbf{0.662} & \textbf{25.423} & \textbf{0.747} \\
    \midrule
    \multirow{7}[0]{*}{\shortstack{\textbf{MSIs}\\\textit{Toy}\\
    \textit{Flowers}\\\textit{Beads}\\\textit{Feathers}\\{\tiny $(256\times 256\times 31)$}\\\textit{PaviaU}\\{\tiny $(256\times 256\times 80)$}}}  & TNN   & 22.808 & 0.555 & 26.199 & 0.714 & 29.950 & 0.837 \\
          & TRLRF & 23.323 & 0.515 & 29.025 & 0.768 & 33.269 & 0.892 \\
          & TCTV  & 28.572 & \underline{0.833} & 32.639 & \underline{0.918} & 36.898 & \underline{0.961} \\
          & HLRTF & 27.758 & 0.759 & 33.442 & 0.912 & \underline{38.839} & 0.970 \\
          & LRTFR & \underline{29.303} & 0.822 & \underline{33.757} & 0.915 & 38.332 & 0.961 \\
          & FLRTF & 21.895 & 0.447 & 29.414 & 0.788 & 36.363 & 0.928 \\
          & GSLR  & \textbf{30.772} & \textbf{0.888} & \textbf{36.333} & \textbf{0.967} & \textbf{41.466} & \textbf{0.988} \\
    \bottomrule
    \end{tabular}%
      \label{tab:random}%
\end{table}%

\begin{table}[!h]
\footnotesize
\setlength{\tabcolsep}{3pt}
\renewcommand\arraystretch{1}
  \caption{The average quantitative results for tube missing. The \textbf{best} and \underline{second-best} values are highlighted.}
    \begin{tabular}{cccccccc}
    \toprule
    \multicolumn{2}{c}{Sampling Rate} & \multicolumn{2}{c}{0.10} & \multicolumn{2}{c}{0.15} & \multicolumn{2}{c}{0.20} \\
    \midrule
    Data  & Method & \multicolumn{1}{c}{PSNR} & \multicolumn{1}{c}{SSIM} & \multicolumn{1}{c}{PSNR} & \multicolumn{1}{c}{SSIM} & \multicolumn{1}{c}{PSNR} & \multicolumn{1}{c}{SSIM} \\
    \midrule
    \multirow{7}[0]{*}{\shortstack{\textbf{Color images}\\\textit{Plane}\\
    \textit{House}\\\textit{Baboon}\\{\tiny $(512\times 512\times 3)$}\\\textit{Female}\\{\tiny $(256\times 256\times 3)$}}}  & TNN   & 18.696 & 0.327 & 20.139 & 0.413 & 21.325 & 0.489 \\
          & TRLRF & 16.912 & 0.211 & 19.127 & 0.359 & 20.861 & 0.470 \\
          & TCTV  & \underline{23.710} & \underline{0.662} & \underline{24.792} & \underline{0.715} & \underline{25.605} & \underline{0.755} \\
          & HLRTF & 20.439 & 0.427 & 21.985 & 0.519 & 23.169 & 0.589 \\
          & LRTFR & 22.126 & 0.548 & 23.550 & 0.642 & 24.828 & 0.711 \\
          & FLRTF & 17.251 & 0.248 & 19.346 & 0.401 & 20.846 & 0.475 \\
          & GSLR  & \textbf{23.999} & \textbf{0.685} & \textbf{25.197} & \textbf{0.737} & \textbf{26.062} & \textbf{0.776} \\
    \midrule
    \multirow{7}[0]{*}{\shortstack{\textbf{MSIs}\\\textit{Toy}\\
    \textit{Flowers}\\\textit{Beads}\\\textit{Feathers}\\{\tiny $(256\times 256 \times 31)$}\\\textit{PaviaU}\\{\tiny $(256\times 256\times 80)$}}}  & TNN   & 21.133 & 0.460 & 22.653 & 0.536 & 23.962 & 0.605 \\
          & TRLRF & 16.896 & 0.127 & 18.826 & 0.322 & 21.271 & 0.436 \\
          & TCTV  & \underline{26.927} & \underline{0.774} & \underline{28.382} & \underline{0.823} & \underline{29.554} & \underline{0.858} \\
          & HLRTF & 23.336 & 0.608 & 25.074 & 0.686 & 26.489 & 0.747 \\
          & LRTFR & 24.176 & 0.604 & 26.580 & 0.729 & 28.065 & 0.804 \\
          & FLRTF & 18.495 & 0.318 & 19.764 & 0.369 & 21.921 & 0.478 \\
          & GSLR  & \textbf{27.359} & \textbf{0.821} & \textbf{29.237} & \textbf{0.869} & \textbf{30.435} & \textbf{0.898} \\
    \bottomrule
    \end{tabular}%
  \label{tab:tube}%
\end{table}%

\begin{table*}[!t]
\footnotesize
\setlength{\tabcolsep}{3.4pt}
\renewcommand\arraystretch{1}
  \centering
  \caption{The quantitative results for slice missing. The \textbf{best} and \underline{second-best} values are highlighted.}
    \begin{tabular}{ccccccccccccccc}
    \toprule
    Method & \multicolumn{2}{c}{TNN} & \multicolumn{2}{c}{TRLRF} & \multicolumn{2}{c}{TCTV} & \multicolumn{2}{c}{HLRTF} & \multicolumn{2}{c}{LRTFR} & \multicolumn{2}{c}{FLRTF} & \multicolumn{2}{c}{GSLR} \\
    \midrule
    Data  & PSNR  & SSIM  & PSNR  & SSIM  & PSNR  & SSIM  & PSNR  & SSIM  & PSNR  & SSIM  & PSNR  & SSIM  & PSNR  & SSIM \\
    \midrule
    \textit{Painting} & 16.217 & 0.107 & 19.954 & 0.430 & 30.556 & 0.886 & 25.997 & 0.661 & 24.834 & 0.716 & \underline{35.833} & \underline{0.949} & \textbf{36.815} & \textbf{0.958} \\
     \textit{Toy} & 11.000 & 0.257 & 9.572 & 0.056 & 27.462 & 0.840 & 27.829 & 0.829 & 24.461 & 0.836 & \underline{33.562} & \underline{0.930} & \textbf{34.201} & \textbf{0.943}  \\
     \textit{Hair}  & 19.887 & 0.294 & 9.006 & 0.070 & 34.757 & 0.892 & 35.033 & 0.902 & 32.228 & 0.853 & \underline{42.563} & \underline{0.973} & \textbf{43.276} & \textbf{0.989} \\
     \textit{Face}  & 17.968 & 0.326 & 7.954 & 0.099 & 31.575 & 0.897 & 30.115 & 0.874 & 22.977 & 0.506 & \underline{36.899} & \underline{0.942} & \textbf{37.246} & \textbf{0.979} \\
    \textit{Cd}    & 18.421 & 0.159 & 19.360 & 0.435 & 23.057 & 0.752 & 22.888 & \underline{0.893} & 22.650 & 0.777 & \underline{23.353} & 0.612 & \textbf{24.507} & \textbf{0.921} \\
    \bottomrule
    \end{tabular}%
  \label{tab:slice}%
\end{table*}%

\begin{figure*}[!h]
	\centering
     \footnotesize
	\setlength{\tabcolsep}{1pt}
	\renewcommand\arraystretch{1}
	\begin{tabular}{cccccccccccc}
  \rotatebox{90}{\makebox[1.6cm]{\textit{Baboon}}} & 
\includegraphics[width=0.1111\textwidth]{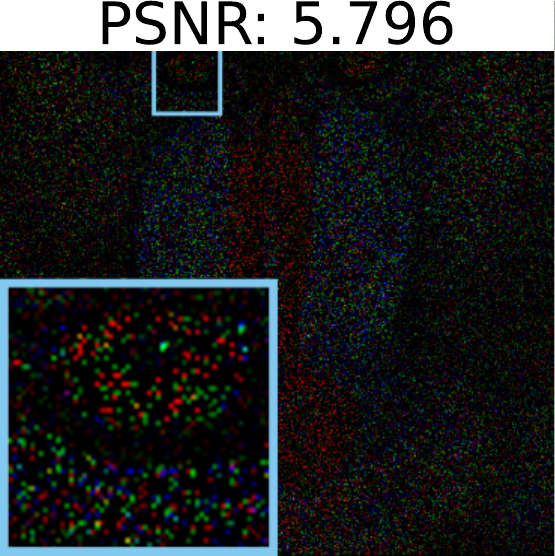}&
\includegraphics[width=0.1111\textwidth]{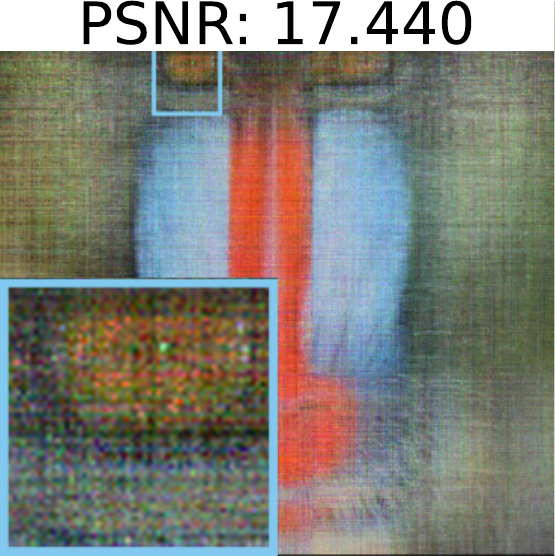}&
\includegraphics[width=0.1111\textwidth]{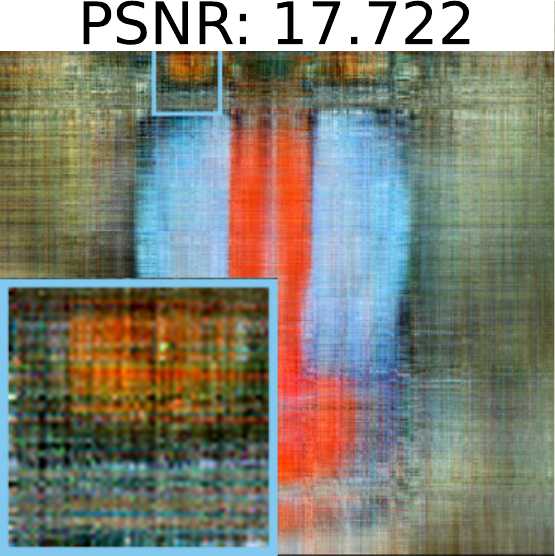}&
\includegraphics[width=0.1111\textwidth]{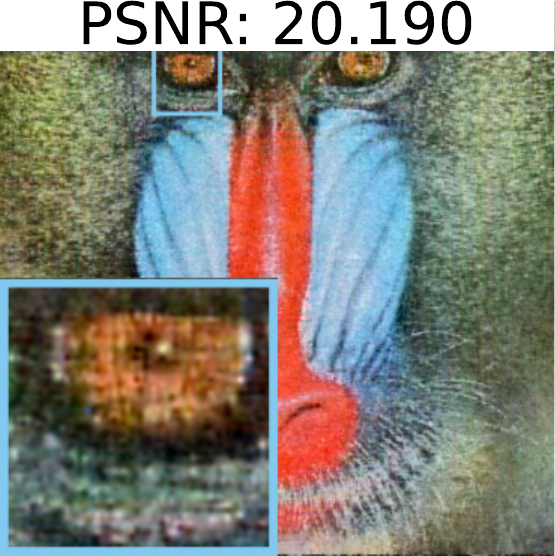}&
\includegraphics[width=0.1111\textwidth]{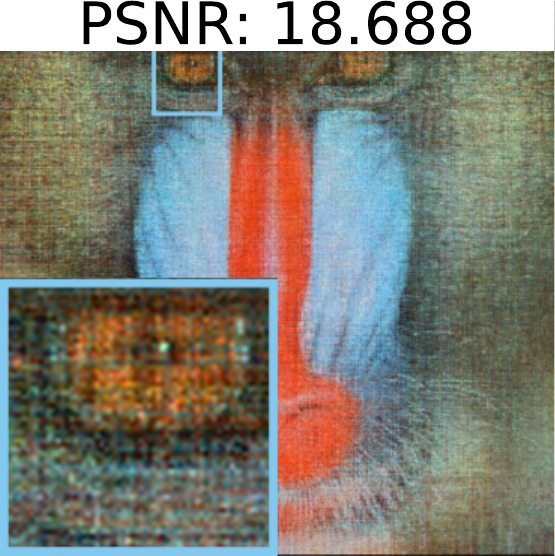}&
\includegraphics[width=0.1111\textwidth]{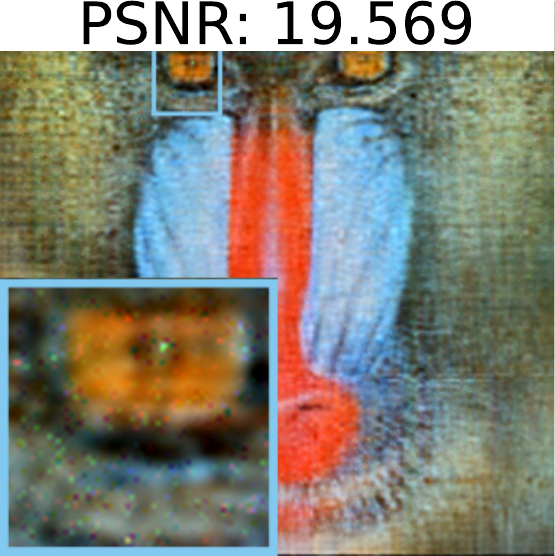}&
\includegraphics[width=0.1111\textwidth]{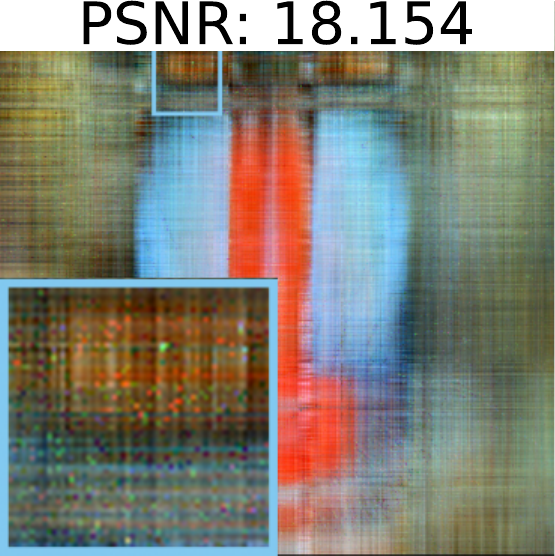}&
\includegraphics[width=0.1111\textwidth]{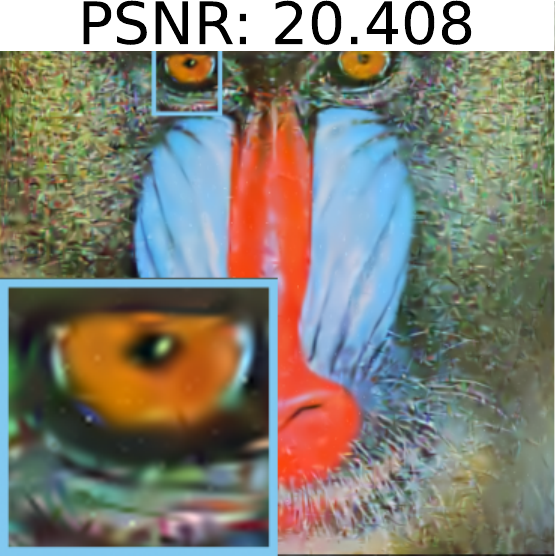}&
\includegraphics[width=0.1111\textwidth]{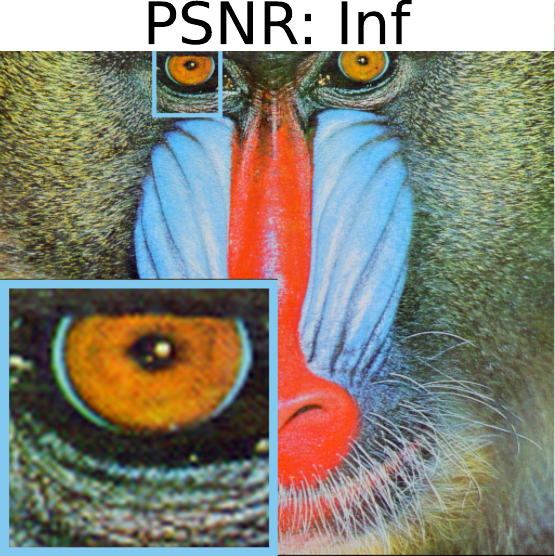}&
\\
\rotatebox{90}{\makebox[1.6cm]{\textit{Feathers}}} & 
\includegraphics[width=0.1111\textwidth]
{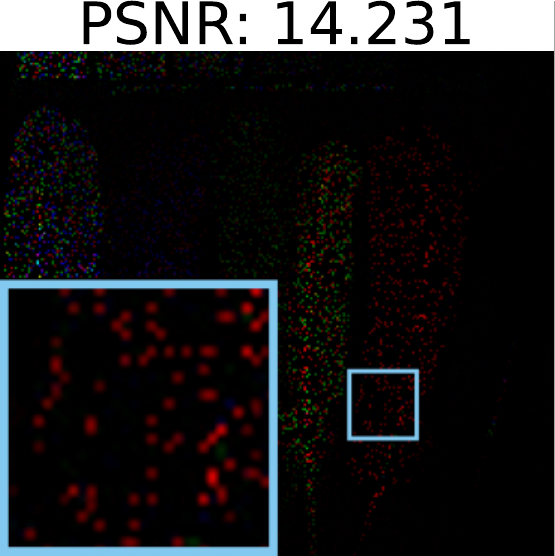}&
\includegraphics[width=0.1111\textwidth]
{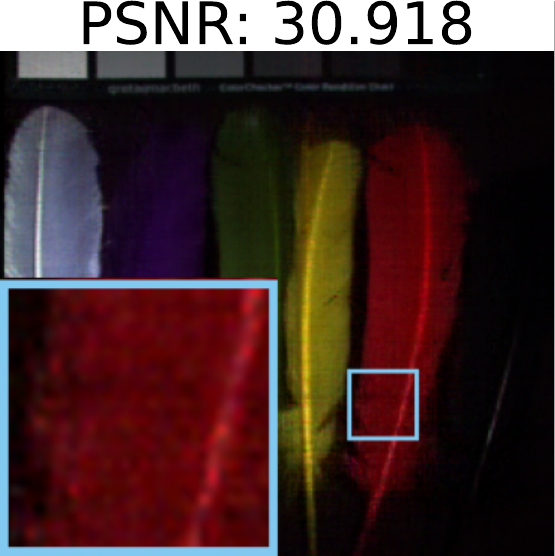}&
\includegraphics[width=0.1111\textwidth]{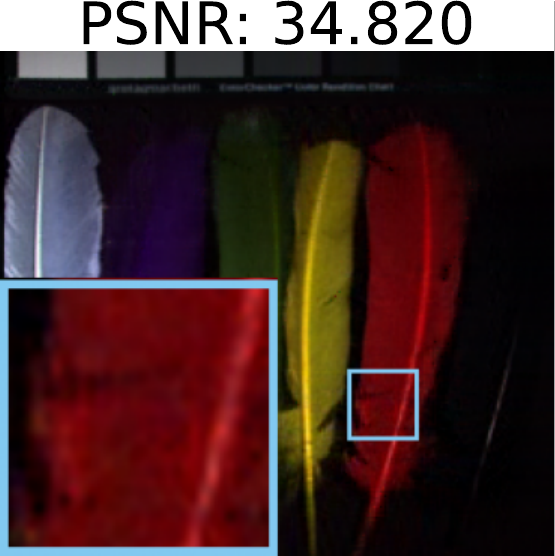}&
\includegraphics[width=0.1111\textwidth]{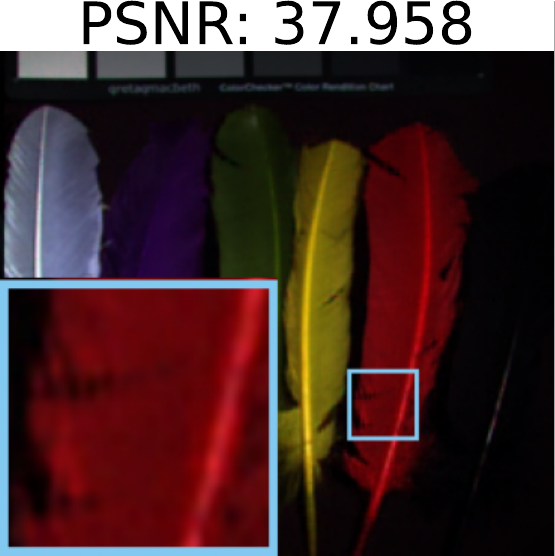}&
\includegraphics[width=0.1111\textwidth]{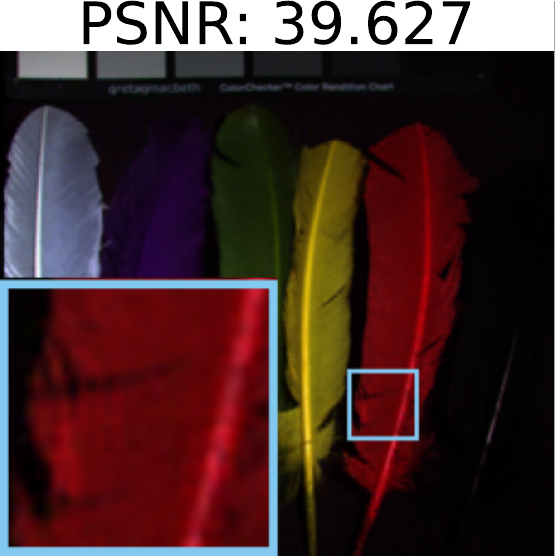}&
\includegraphics[width=0.1111\textwidth]{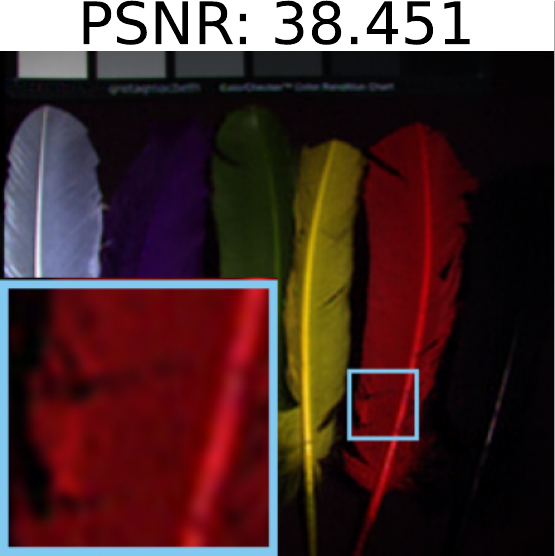}&
\includegraphics[width=0.1111\textwidth]{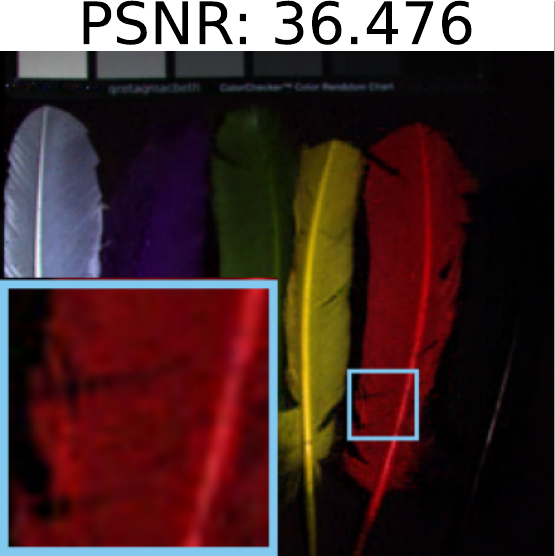}&
\includegraphics[width=0.1111\textwidth]{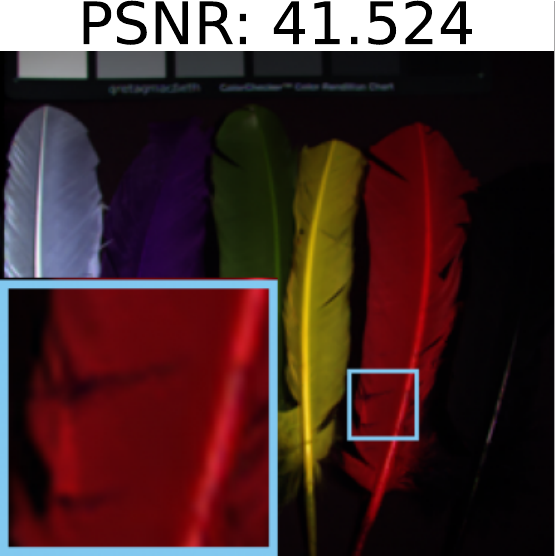}&
\includegraphics[width=0.1111\textwidth]{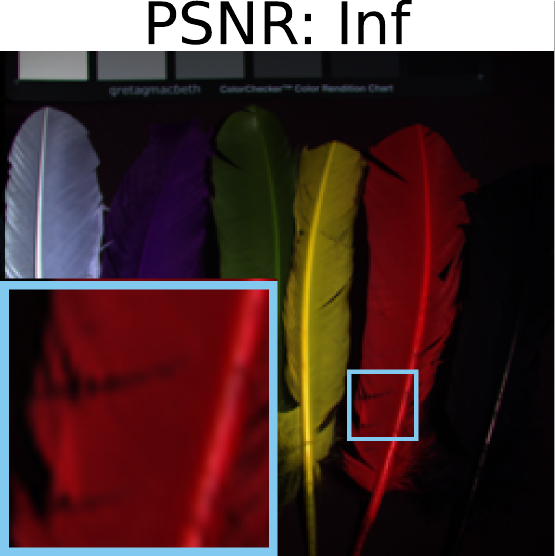}&
\\
&\textbf{Observed} & \textbf{TNN} & \textbf{TRLRF} & \textbf{TCTV} & \textbf{HLRTF} & \textbf{LRTFR} &\textbf{FLRTF}& \textbf{GSLR} & \textbf{GT} 
\end{tabular}
\caption{Reconstructed results and zoomed-in details by different methods under the random missing (SR = 0.10).}
\label{fig:random}
 \end{figure*}
 
\begin{figure*}[!h]
	\centering
     \footnotesize
	\setlength{\tabcolsep}{1pt}
	\renewcommand\arraystretch{1}
\begin{tabular}{cccccccccccc}
  \rotatebox{90}{\makebox[1.6cm]{\textit{Plane}}} & 
\includegraphics[width=0.1111\textwidth]{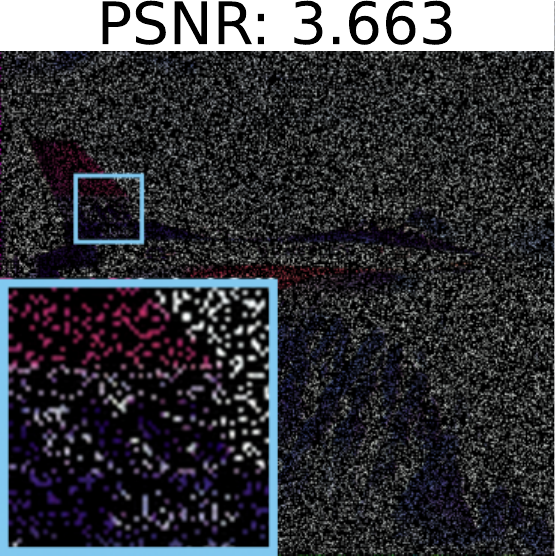}&
\includegraphics[width=0.1111\textwidth]{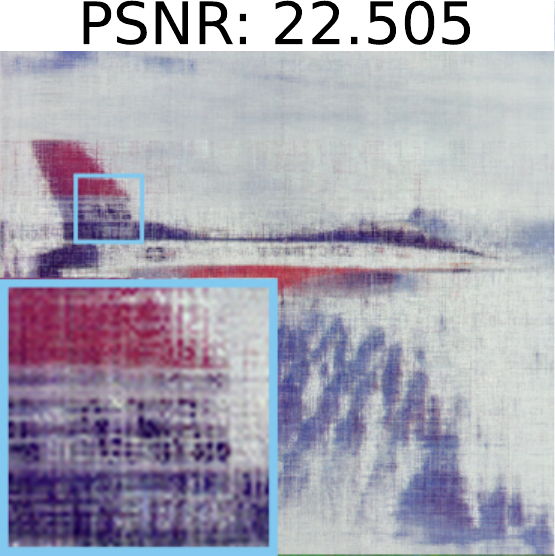}&
\includegraphics[width=0.1111\textwidth]{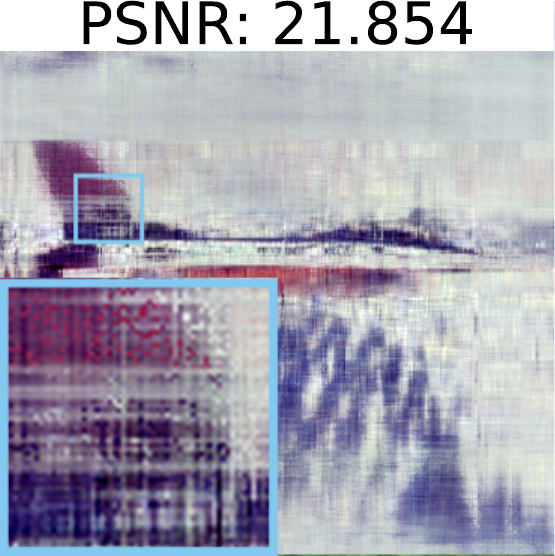}&
\includegraphics[width=0.1111\textwidth]{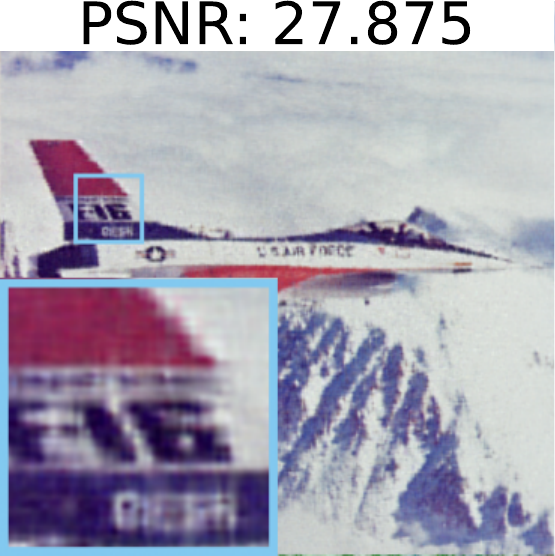}&
\includegraphics[width=0.1111\textwidth]{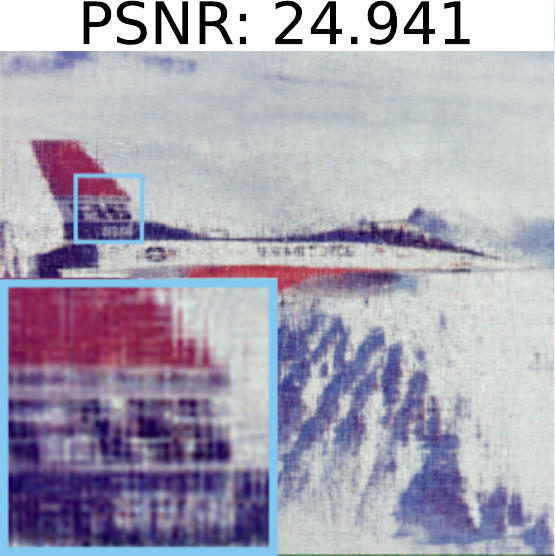}&
\includegraphics[width=0.1111\textwidth]{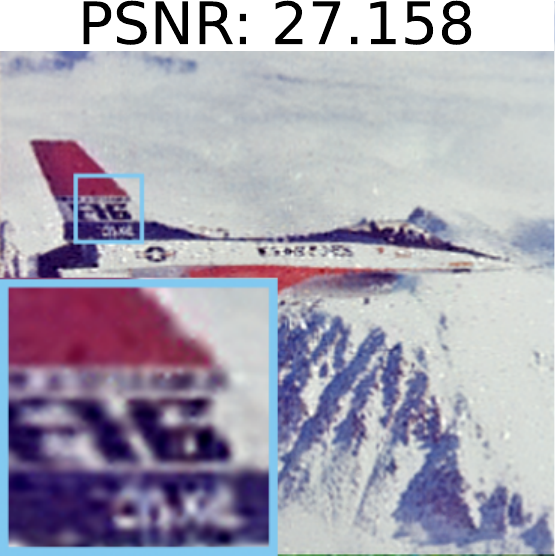}&
\includegraphics[width=0.1111\textwidth]{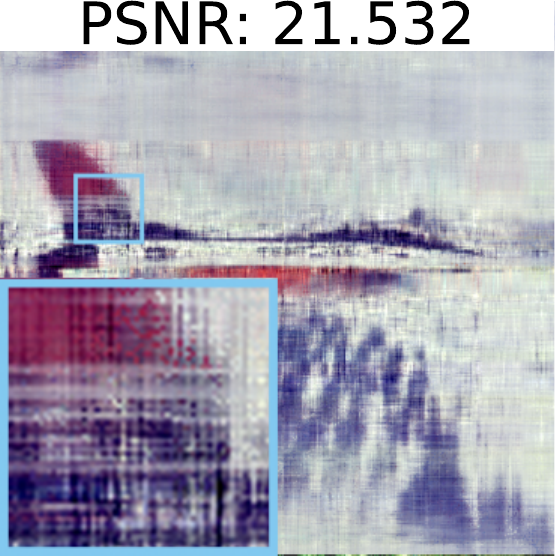}&
\includegraphics[width=0.1111\textwidth]{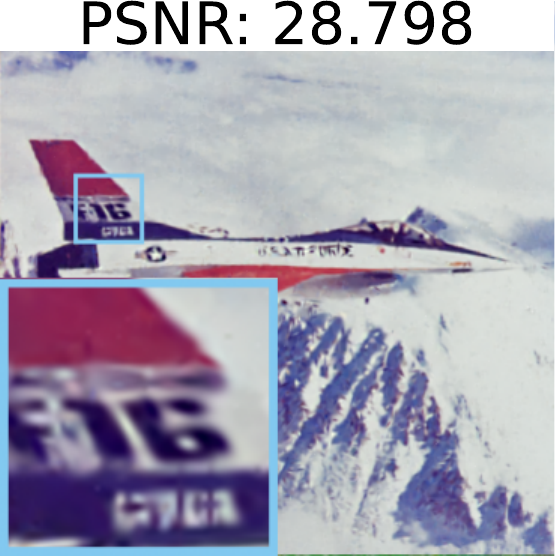}&
\includegraphics[width=0.1111\textwidth]{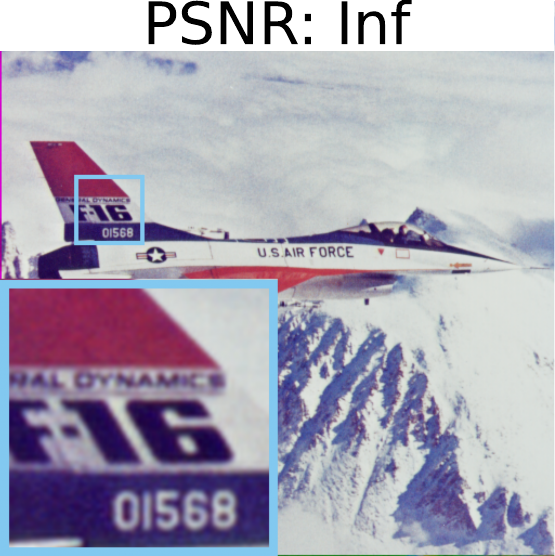}&
\\
\rotatebox{90}{\makebox[1.6cm]{\textit{PaviaU}}} & 
\includegraphics[width=0.1111\textwidth]{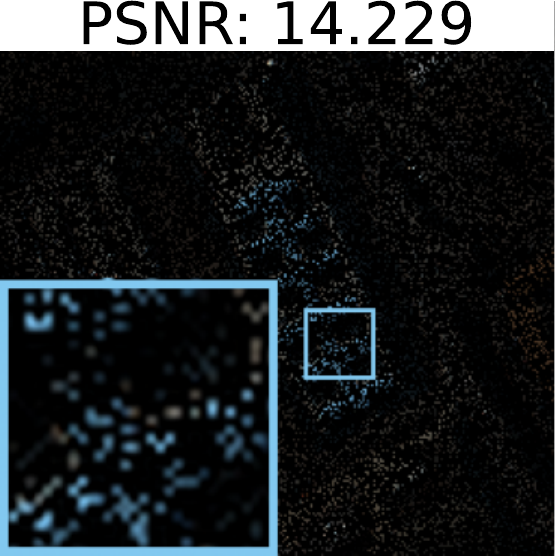}&
\includegraphics[width=0.1111\textwidth]{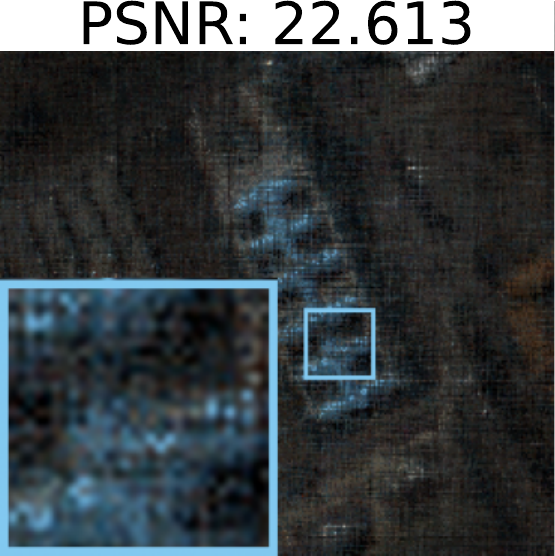}&
\includegraphics[width=0.1111\textwidth]{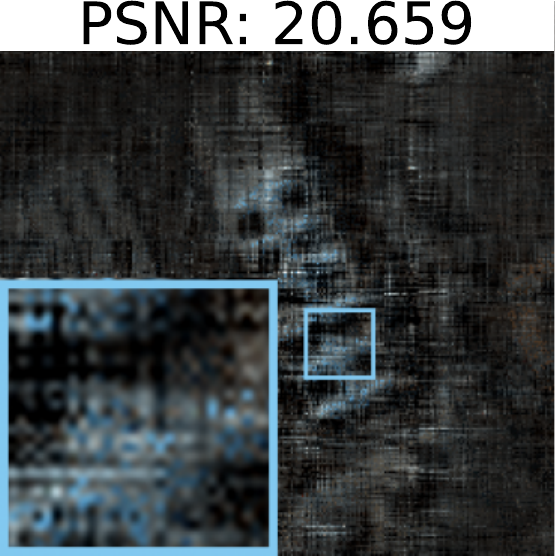}&
\includegraphics[width=0.1111\textwidth]{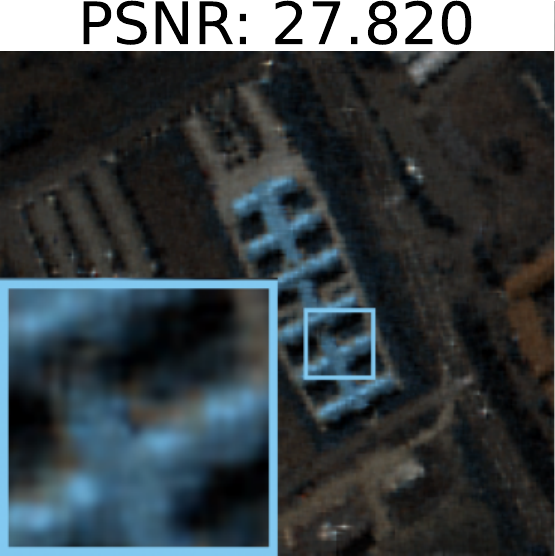}&
\includegraphics[width=0.1111\textwidth]{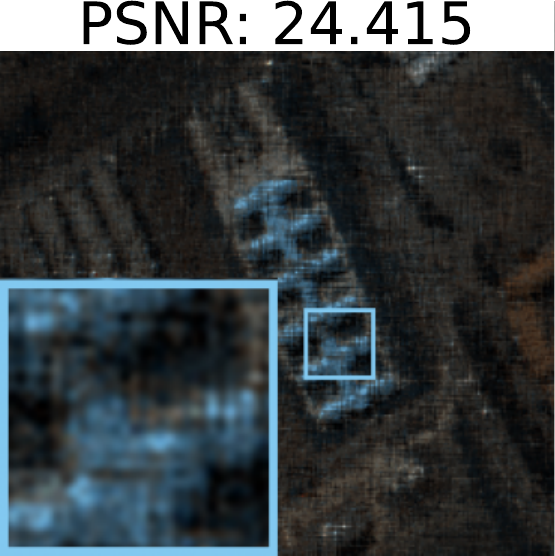}&
\includegraphics[width=0.1111\textwidth]{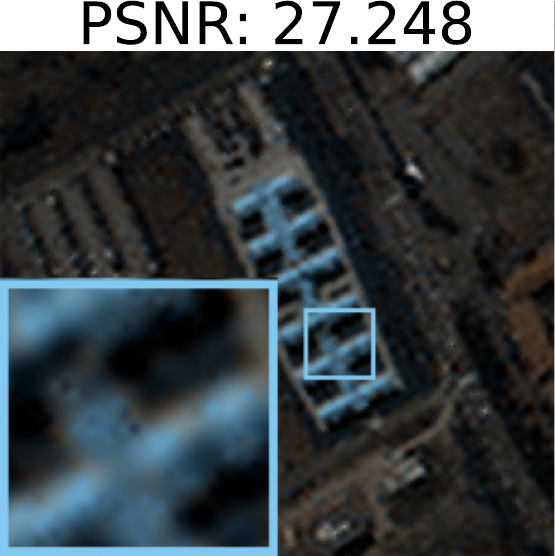}&
\includegraphics[width=0.1111\textwidth]{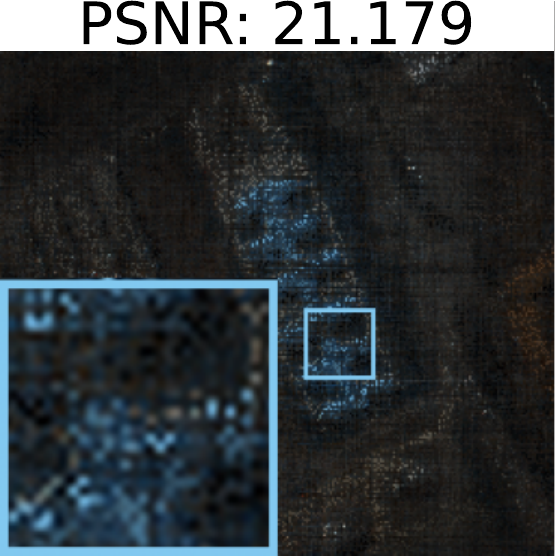}&
\includegraphics[width=0.1111\textwidth]{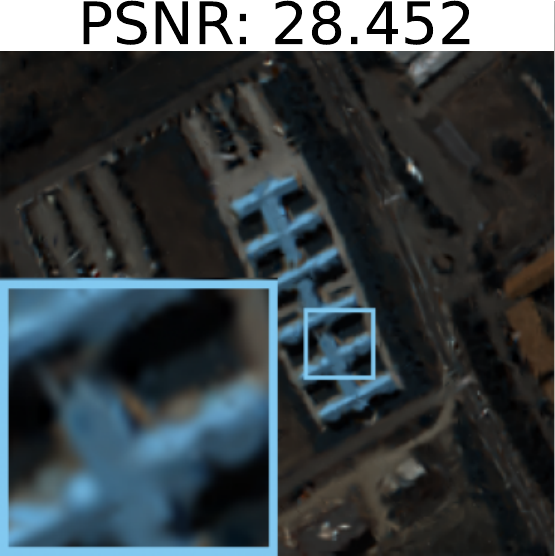}&
\includegraphics[width=0.1111\textwidth]{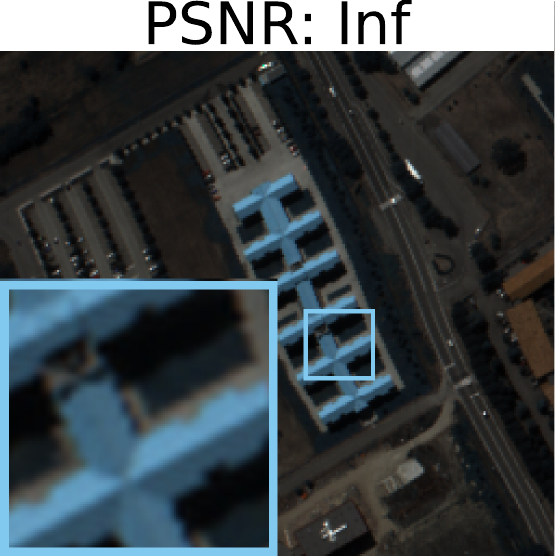}&
\\
&\textbf{Observed} & \textbf{TNN} & \textbf{TRLRF} & \textbf{TCTV} & \textbf{HLRTF} & \textbf{LRTFR} &\textbf{FLRTF}& \textbf{GSLR} & \textbf{GT} 
\end{tabular}
\caption{Reconstructed results and zoomed-in details by different methods under the tube missing (SR = 0.20).}
\label{fig:tube}

 \end{figure*}

 \begin{figure*}[!h]
	\centering
     \footnotesize
	\setlength{\tabcolsep}{1pt}
	\renewcommand\arraystretch{0.5}
    \begin{tabular}{cccccccccc}
\rotatebox{90}{\makebox[1.6cm]{\textit{Painting}}} & 
\includegraphics[width=0.1111\textwidth]{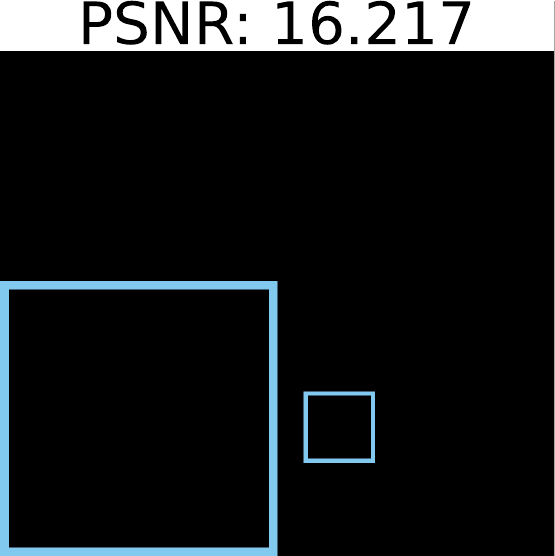}&
\includegraphics[width=0.1111\textwidth]{imgs/zoom_1/painting_e5_slicemiss/TNN.pdf}&
\includegraphics[width=0.1111\textwidth]{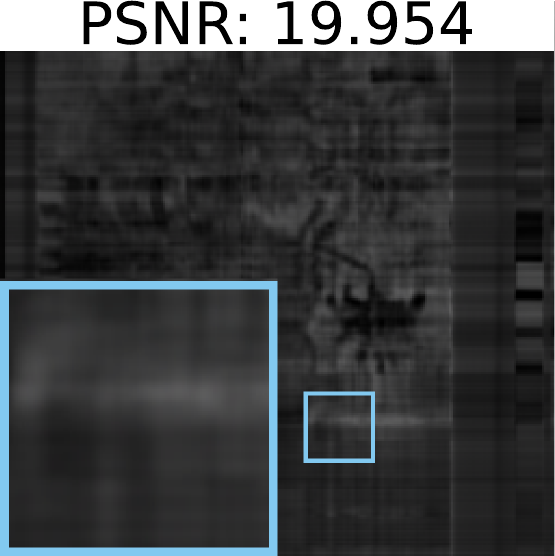}&
\includegraphics[width=0.1111\textwidth]{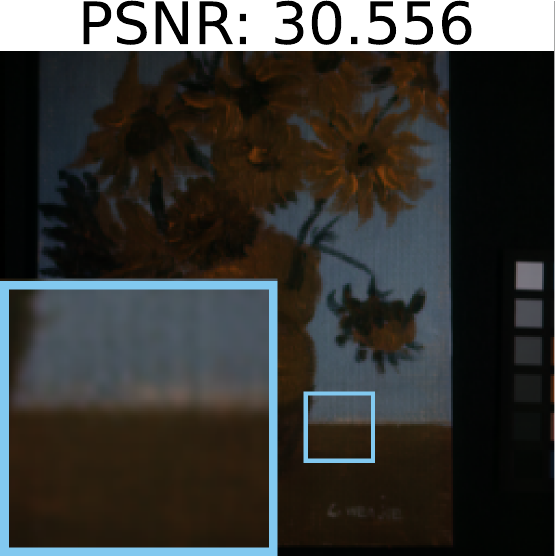}&
\includegraphics[width=0.1111\textwidth]{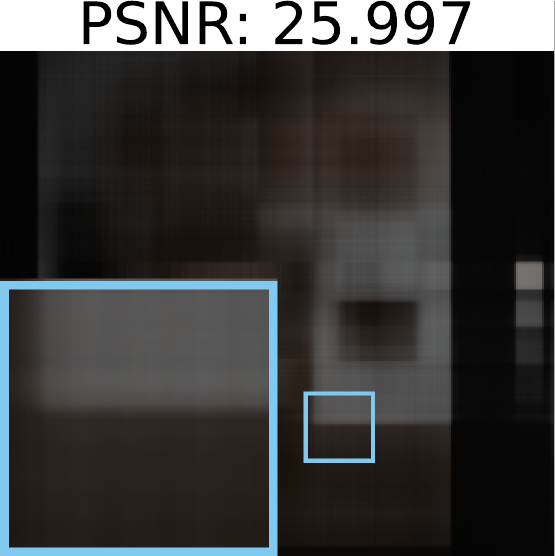}&
\includegraphics[width=0.1111\textwidth]{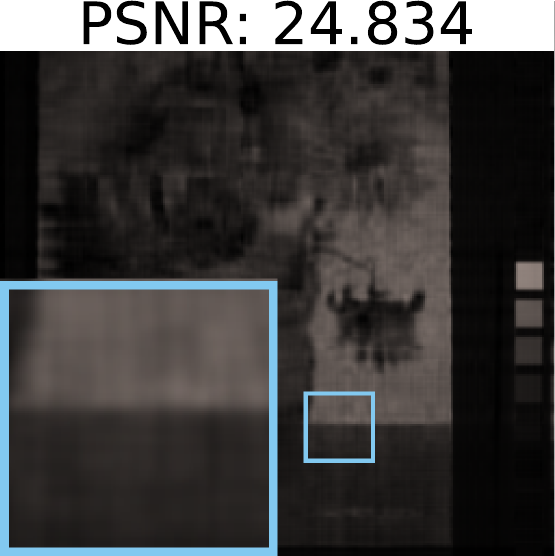}&
\includegraphics[width=0.1111\textwidth]{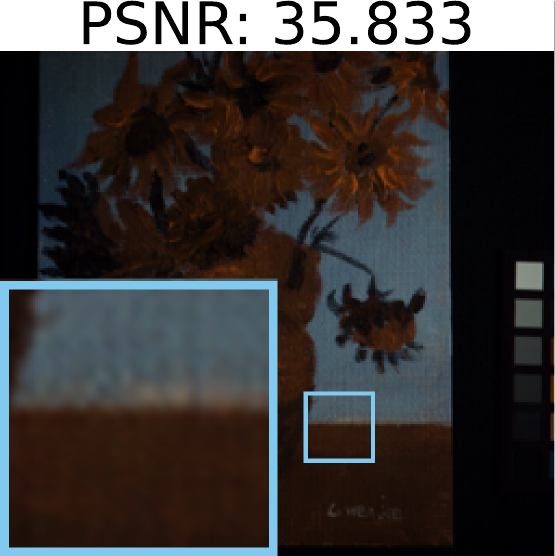}&
\includegraphics[width=0.1111\textwidth]{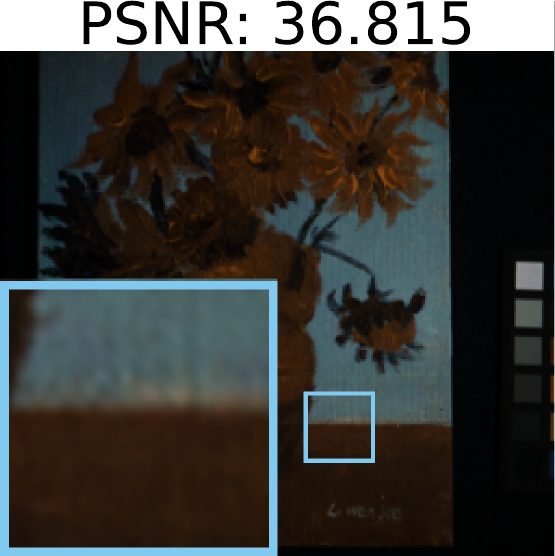}&
\includegraphics[width=0.1111\textwidth]{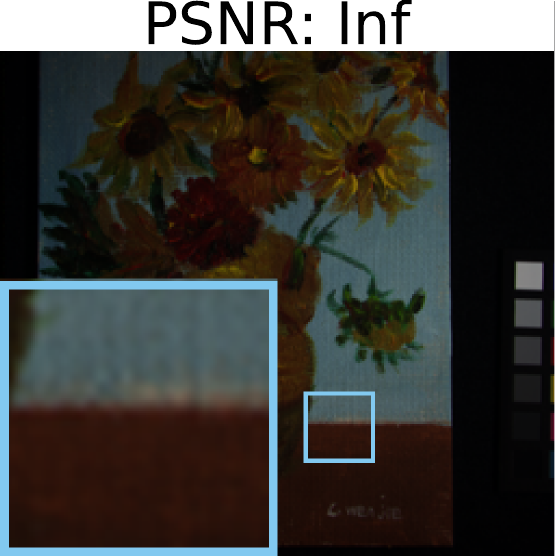}
\\
\rotatebox{90}{\makebox[1.6cm]{\textit{Hairs}}} & 
\includegraphics[width=0.1111\textwidth]{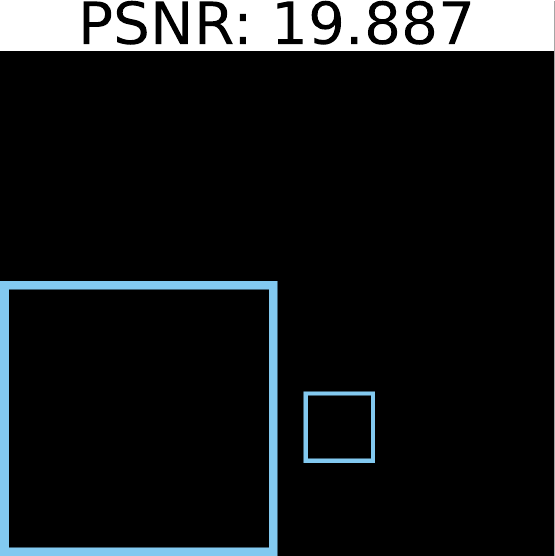}&
\includegraphics[width=0.1111\textwidth]{imgs/zoom_1/hairs_e5_slicemiss/TNN.pdf}&
\includegraphics[width=0.1111\textwidth]{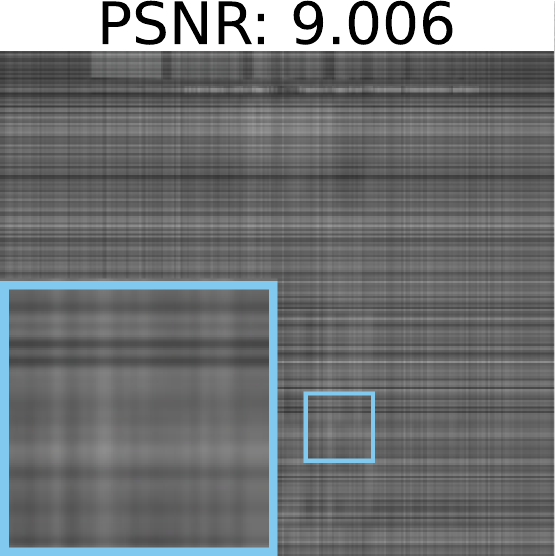}&
\includegraphics[width=0.1111\textwidth]{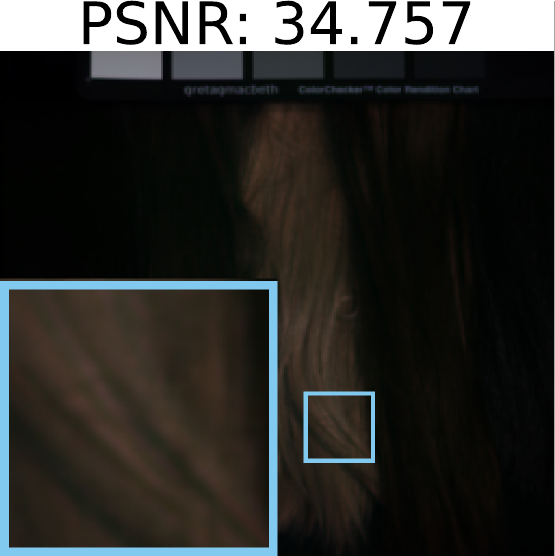}&
\includegraphics[width=0.1111\textwidth]{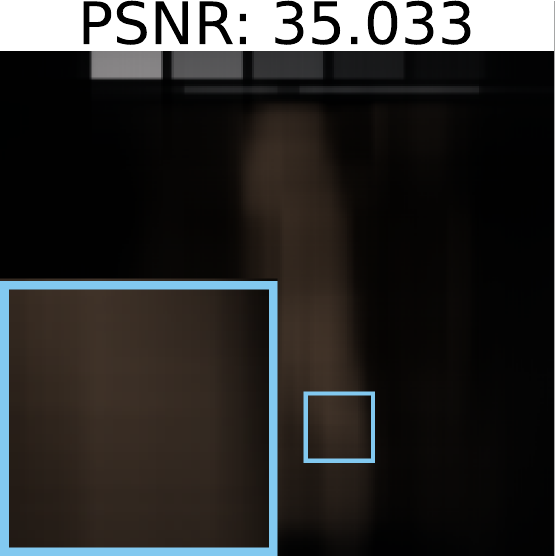}&
\includegraphics[width=0.1111\textwidth]{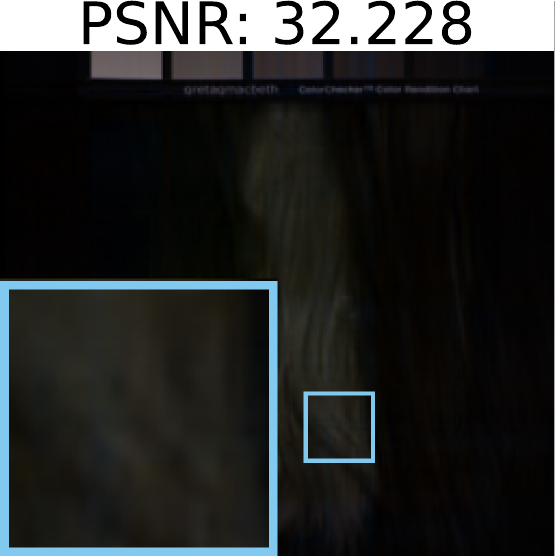}&
\includegraphics[width=0.1111\textwidth]{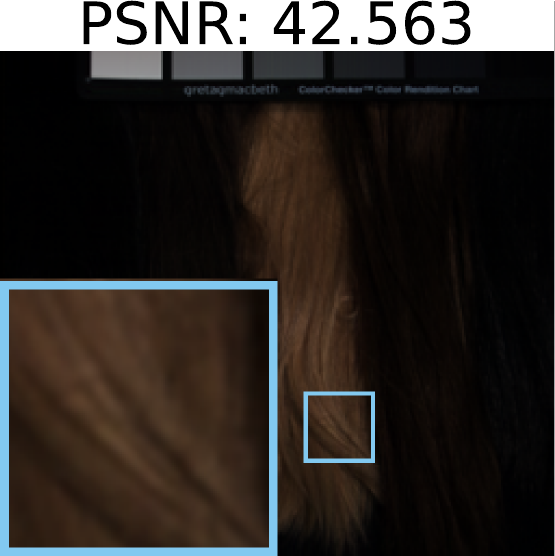}&
\includegraphics[width=0.1111\textwidth]{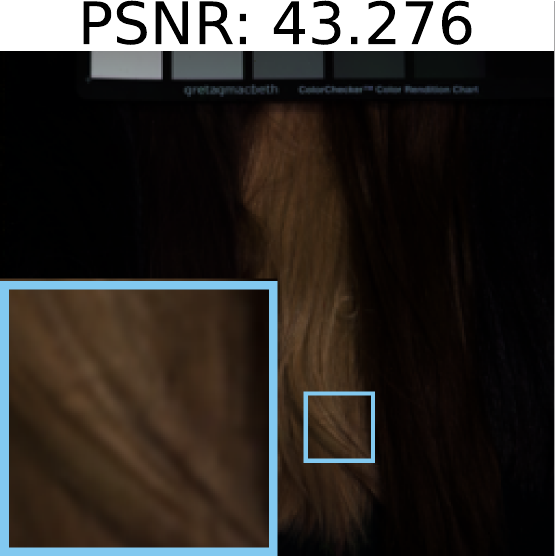}&
\includegraphics[width=0.1111\textwidth]{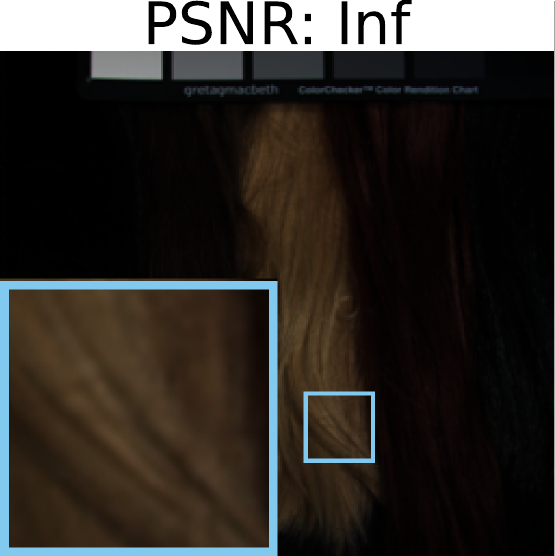}
\\
&\textbf{Observed} & \textbf{TNN} & \textbf{TRLRF} & \textbf{TCTV} & \textbf{HLRTF} & \textbf{LRTFR} &  \textbf{FLRTF} & \textbf{GSLR} & \textbf{GT} 
\end{tabular}
\caption{Reconstructed results and zoomed-in details by different methods under the slice missing.}
\label{fig:slice}
 \end{figure*}

\subsection{Experimental Results}
The quantitative results of multi-dimensional image recovery (\ie, random missing, tube missing, and slice missing) are respectively reported in \cref{tab:random,tab:tube,tab:slice}.
It can be observed that the proposed GSLR achieves the highest PSNR and SSIM values, demonstrating its superior overall reconstruction quality and structural consistency even under extremely low sampling rates.

To facilitate visual comparison, \cref{fig:random,fig:tube,fig:slice} present several recovered pseudo-color images. As shown in \cref{fig:random,fig:tube}, GSLR successfully reconstructs multi-dimensional images and captures spatial high-frequency details (\eg, the eye region of the Baboon and the rooftop structures in the House) more accurately than previous t-SVD methods, benefiting from its powerful representation capability. In \cref{fig:slice}, it is evident that TNN and TRLRF exhibit unsatisfactory performance, as their transforms rely on fixed atoms and ignore the continuity along the mode-3 fibers. Although HLRTF and LRTFR recover most missing information, they still suffer from spatial and spectral distortions. TCTV and FLRTF produce relatively better results but remain inadequate in capturing the spectral information. In contrast, GSLR produces recovered results closest to GT, highlighting the strong capability in capturing high-frequency information along the mode-3 fibers.

\section{Discussions}

\subsection{The Contribution of 2D Gaussian Splatting-based Latent Tensor Approximation}
To assess the contribution of 2D Gaussian splatting-based latent tensor (2DGS-LT) approximation, we replace it with two alternatives: an unconstrained approximation (\ie, a parameterized latent tensor) and a low-rank factorization \cite{HLRTF}. The PSNR and SSIM values recovered from different approximations of the latent tensor are shown in \cref{tab:ablation_c}. It is obvious that selecting 2DGS-LT outperforms both the unconstrained approximation and low-rank factorization. Furthermore, \cref{fig:2dgs} presents a visual comparison of the recovered result of different approximations of the latent tensor. It is evident that 2DGS-LT provides superior characterization of spatial local high-frequency information.

\begin{table}[!h]
  \centering
    \footnotesize
  \setlength{\tabcolsep}{3pt}
\renewcommand\arraystretch{1}
  \caption{Quantitative results of different approximations of the latent tensor on the \textit{Toy} dataset.}
    \begin{tabular}{cccccccc}
    \toprule
    \multicolumn{2}{c}{\begin{tabular}{@{}c@{}}  \textbf{\textit{Latent Tensor}}\\  \textbf{\textit{Approximation}} \end{tabular}} & \multicolumn{2}{c}{Unconstrained} & \multicolumn{2}{c}{\begin{tabular}{@{}c@{}}Low-rank \\ Factorization\end{tabular}} & \multicolumn{2}{c}{2DGS-LT} \\
    \midrule
    Task  & SR   & PSNR  & SSIM  & PSNR  & SSIM  & PSNR  & SSIM \\
    \midrule
    \multirow{3}[0]{*}{Random} & 0.02  & 11.546 & 0.045 & 25.588 & 0.725 &  \textbf{31.712} & \textbf{0.923} \\
          & 0.05 &  16.813 & 0.197 & 31.968& 0.906 & \textbf{37.148} & \textbf{0.983} \\
          & 0.10 & 36.875 & 0.963 & 37.955& 0.963 & \textbf{43.630} & \textbf{0.995} \\
    \midrule
    \multirow{3}[0]{*}{Tube} & 0.10  & 20.864 & 0.651 & 19.352 & 0.412 & \textbf{27.713} &  \textbf{0.892} \\
          & 0.15 & 23.144 & 0.719 & 21.720 & 0.543 &  \textbf{29.347} & \textbf{0.911} \\
          & 0.20 & 24.903 & 0.773 &23.240 & 0.647& \textbf{30.654} & \textbf{0.938} \\
          \midrule
    Slice &  - & 28.123 & 0.823 & 30.323 & 0.853&\textbf{34.201} & \textbf{0.943} \\
    \bottomrule
    \end{tabular}%
  \label{tab:ablation_c}
\end{table}%

\begin{figure}[!h]
    \centering
    \footnotesize
    \setlength{\tabcolsep}{0.8pt}
    \renewcommand\arraystretch{1}
    \begin{tabular}{cccc}
        \includegraphics[width=0.25\columnwidth]{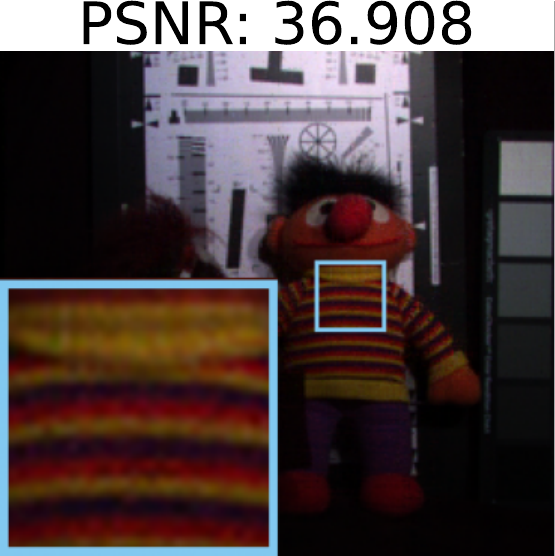} &
        \includegraphics[width=0.25\columnwidth]{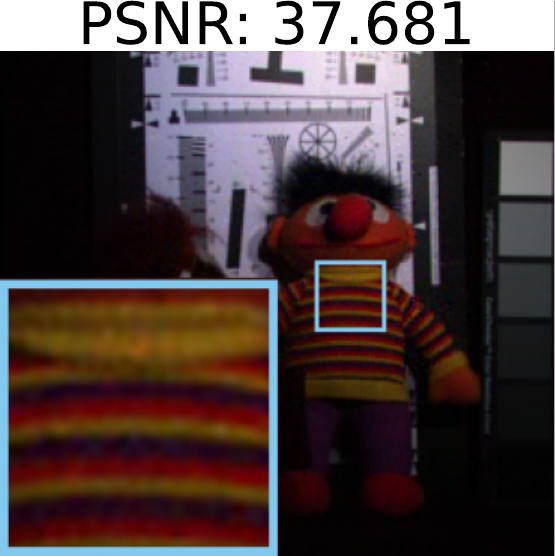} &
        \includegraphics[width=0.25\columnwidth]{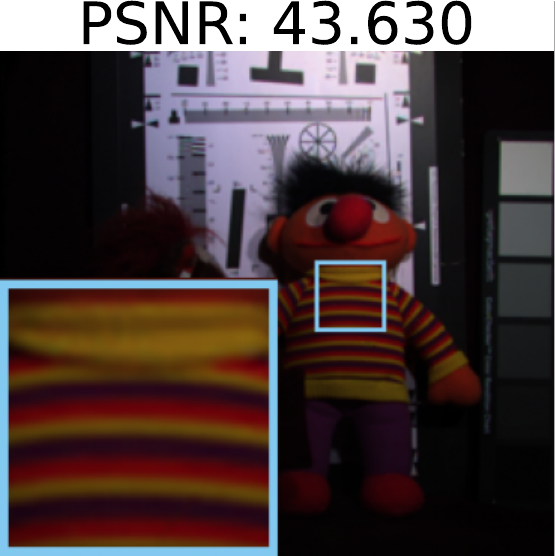} &
        \includegraphics[width=0.25\columnwidth]{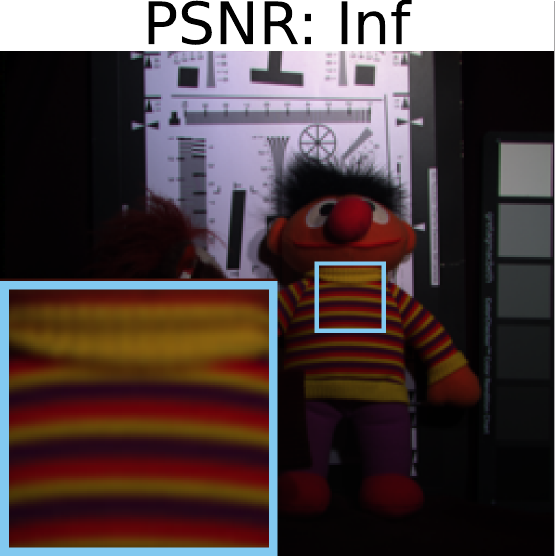} \\
        (a) & (b) & (c) & (d)
    \end{tabular}
    \caption{Influence of different approximations of the latent tensor on the \textit{Toy} dataset under the random missing (SR = 0.10). (a) Unconstrained approximation, (b) Low-rank factorization, (c) 2DGS-LT, and (d) GT.}
  \label{fig:2dgs}
\end{figure}

\begin{table}[!h]
  \centering
    \footnotesize
  \setlength{\tabcolsep}{3pt}
\renewcommand\arraystretch{1}
  \caption{Quantitative results of different transforms on the \textit{Toy} dataset.}
    \begin{tabular}{cccccccc}
    \toprule
    \multicolumn{2}{c}{\begin{tabular}{@{}c@{}}  \textbf{\textit{Transform Matrix}}\\  \textbf{\textit{Approximation}}\end{tabular}} & \multicolumn{2}{c}{Unconstrained} & \multicolumn{2}{c}{INR} & \multicolumn{2}{c}{1DGS-TM} \\
    \midrule
    Task  & SR   & PSNR  & SSIM  & PSNR  & SSIM  & PSNR  & SSIM \\
    \midrule
    \multirow{3}[0]{*}{Random} & 0.02  & 26.791 & 0.782 &27.888&	0.841 & \textbf{31.712} & \textbf{0.923} \\
          & 0.05 & 31.181 & 0.930&33.760& 0.935 & \textbf{37.148} & \textbf{0.983} \\
          & 0.10 & 38.692 & 0.985& 39.330 & 0.973 & \textbf{43.630} & \textbf{0.995} \\
    \midrule
    \multirow{3}[0]{*}{Tube} & 0.10  &  27.130 & 0.872 &22.158 & 0.604 & \textbf{27.318} &  \textbf{0.868} \\
          & 0.15 & 29.120 & 0.912 &24.867& 0.669&  \textbf{29.347} & \textbf{0.911} \\
          & 0.20 &  29.881 & 0.903&25.850& 0.740 & \textbf{30.654} & \textbf{0.938} \\
              \midrule
    Slice & -  & 11.067 & 0.224& 26.637 &	0.786  &\textbf{34.201} & \textbf{0.943} \\
    \bottomrule
    \end{tabular}%
  \label{tab:ablation_t}
\end{table}%

\subsection{The Contribution of 1D Gaussian Splatting-based Transform Matrix Approximation}

\begin{figure}[t]
  \centering
  \includegraphics[width=.5\columnwidth]{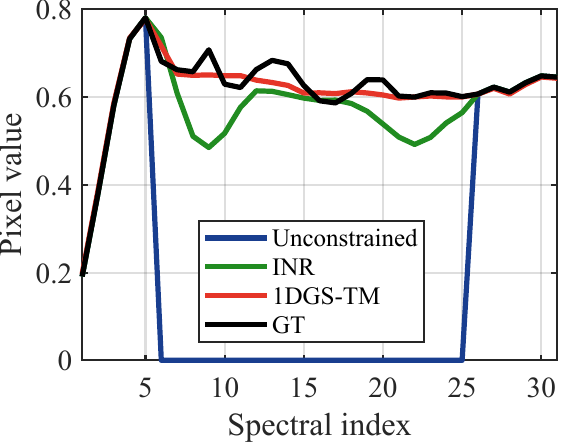}
  \caption{The visualization of recovered spectral curves with different transforms  at the spatial location $(50, 90)$ on the \textit{Toy} dataset under the slice missing (\ie, Unconstrained transform, INR, and 1DGS-TM).}
    \label{fig:1dgs}
\end{figure}

To analyze the contribution of 1D Gaussian splatting-based transform matrix (1DGS-TM) approximation, we replace it with two alternatives: an unconstrained transform (\ie, a parameterized transform matrix) and implicit neural representation (INR) \cite{INR}-based transform matrix. We compare the image recovery performance of different transforms. As shown in \cref{tab:ablation_t}, we observe that the 1DGS-TM outperforms its alternatives across three missing patterns. 
Furthermore, we plot the recovered spectral curves in \cref{fig:1dgs}. The unconstrained transform is essentially discrete and suffers from gradient vanishing, which makes it incapable of handling slice missing. Both INR and 1DGS-TM provide continuous spectral characterization, while 1DGS-TM achieves more accurate spectral recovery.

\begin{table}[h]
  \centering
    \footnotesize
  \setlength{\tabcolsep}{3pt}
\renewcommand\arraystretch{.8}
  \caption{Quantitative results of vanilla 2DGS and proposed GSLR on the \textit{Plane} and \textit{Toy} datasets.}
    \begin{tabular}{ccccccc}
    \toprule
    \multicolumn{3}{c}{Method} & \multicolumn{2}{c}{2DGS} & \multicolumn{2}{c}{GSLR} \\
    \midrule
    Data  & Task  & SR    & PSNR  & SSIM  & PSNR  & SSIM \\
    \midrule
    \multirow{6}[0]{*}{\shortstack{\textbf{Color images}\\\textit{Plane}\\{\tiny $(512\times 512\times 3)$}}} & \multirow{3}[2]{*}{Random} & 0.02  & 18.648 & 0.464 & \textbf{23.621} & \textbf{0.736} \\
          &       & 0.05  & 22.127 & 0.638 & \textbf{26.030} & \textbf{0.809} \\
          &       & 0.10  & 23.543 & 0.712 & \textbf{28.304} & \textbf{0.869} \\
\cmidrule{2-7}          & \multirow{3}[2]{*}{Tube} & 0.10  & 22.072 & 0.609 & \textbf{26.124} & \textbf{0.814} \\
          &       & 0.15  & 24.779 & 0.717 & \textbf{27.710} & \textbf{0.853} \\
          &       & 0.20  & 26.498 & 0.792 & \textbf{28.798} & \textbf{0.874} \\
    \midrule
     \multirow{6}[0]{*}{\shortstack{\textbf{MSI}\\\textit{Toy}\\{\tiny $(256\times 256\times 31)$}}} & \multirow{3}[2]{*}{Random} & 0.02  & 22.578 & 0.624 & \textbf{31.712} & \textbf{0.923} \\
          &       & 0.05  & 26.047 & 0.547 & \textbf{37.148} & \textbf{0.983} \\
          &       & 0.10  & 28.770 & 0.927 & \textbf{43.630} & \textbf{0.995} \\
\cmidrule{2-7}          & \multirow{3}[2]{*}{Tube} & 0.10  & 24.818 & 0.734 & \textbf{27.713} & \textbf{0.892} \\
          &       & 0.15  & 26.373 & 0.734 & \textbf{29.347} & \textbf{0.911} \\
          &       & 0.20  & 29.447 & 0.932 & \textbf{30.654} & \textbf{0.938} \\
    \bottomrule
    \end{tabular}%
  \label{tab:GSLRvs2dGS}%
\end{table}%

\subsection{Vanilla 2D Gaussian Splatting \textit{VS} GSLR}
Gaussian splatting provides a new paradigm for continuous and powerful image representation. To compare the performance of the proposed GSLR against the vanilla 2D Gaussian Splatting (2DGS) \cite{GaussianImage}, we report the recovery results of both methods in \cref{tab:GSLRvs2dGS}.

As shown in \cref{tab:GSLRvs2dGS}, we can observe that GSLR outperforms 2DGS on both color images and MSIs under three missing patterns. This phenomenon arises as 2DGS overlooks the inherent structure of multi-dimensional images, such as low-rankness, which is effectively captured by the proposed GSLR. Another important remark is that 2DGS fails to handle slice missing due to its discrete feature attribute, which cannot capture the continuity along the mode-3 fibers.
In contrast, GSLR effectively addresses this limitation by leveraging tailored 1D Gaussian Splatting.

Due to space limitations, more experimental results and discussions (\eg, hyperparameter analysis, detailed result tables and reconstructed images) are provided in the supplementary materials.

\section{Conclusion}

This paper addresses the limitations of the existing t-SVD in capturing local high-frequency information in multi-dimensional images. Specifically, we have proposed a novel framework based on tailored Gaussian splatting (\ie, GSLR) for multi-dimensional image representation. GSLR exhibits strong representation ability, particularly in capturing high-frequency details. Extensive experiments demonstrate that GSLR outperforms state-of-the-art methods.

{
    \small
    \bibliographystyle{ieeenat_fullname}
    \bibliography{main}
}


\end{document}